\documentclass{article}

\PassOptionsToPackage{numbers,compress}{natbib}

\usepackage[preprint]{neurips_2026}

\usepackage[utf8]{inputenc} 
\usepackage[T1]{fontenc}    
\usepackage{hyperref}       
\hypersetup{
  hidelinks,
  pdftitle={Uncertainty Gates the Gradient: Adaptive Updates for Generalized Few-Shot 3D Segmentation},
  pdfauthor={Yifei Zhao, Fanyu Zhao, Shengtang Wu, Han Zhang, Ruike Cao, Yinsheng Li}
}
\usepackage{url}            
\usepackage{booktabs}       
\usepackage{amsfonts}       
\usepackage{nicefrac}       
\usepackage[nopatch=footnote]{microtype}      
\usepackage[table]{xcolor}  
\usepackage{colortbl}

\usepackage{graphicx}
\usepackage{wrapfig}
\usepackage{subcaption}
\usepackage{float}
\usepackage{multirow} 
\usepackage{array}
\usepackage{tikz}
\usetikzlibrary{calc}

\usepackage{amsmath}
\usepackage{amssymb}
\usepackage{mathtools}
\usepackage{amsthm}
\usepackage[capitalize,noabbrev]{cleveref}
\usepackage{enumitem}

\theoremstyle{plain}

\theoremstyle{definition}

\theoremstyle{remark}

\newcommand{\best}[1]{\textbf{#1}}
\newcommand{\second}[1]{\underline{#1}}
\newcommand{\third}[1]{#1}

\usepackage[disable,textsize=tiny]{todonotes}
\title{Uncertainty Gates the Gradient: Adaptive Updates for Generalized Few-Shot 3D Segmentation}

\author{%
  \textbf{Yifei Zhao\textsuperscript{1,*,$\dagger$} \quad Fanyu Zhao\textsuperscript{1,*} \quad Shengtang Wu\textsuperscript{1}} \\
  \textbf{Han Zhang\textsuperscript{1} \quad Ruike Cao\textsuperscript{2} \quad Yinsheng Li\textsuperscript{1,$\dagger$}} \\[0.5em]
  \normalfont\textsuperscript{1}Fudan University \\
  \normalfont\textsuperscript{2}University of Science and Technology of China
}

\makeatletter
\renewcommand{\@notice}{\enlargethispage{2\baselineskip}}
\makeatother

\begin{document}

\maketitle

\begingroup
\renewcommand{\thefootnote}{\fnsymbol{footnote}}
\footnotetext[1]{Yifei Zhao and Fanyu Zhao contributed equally to this work.\quad\textit{Preprint.}}
\footnotetext[2]{Corresponding authors: Yifei Zhao (\href{mailto:yfzhao19@fudan.edu.cn}{\texttt{yfzhao19@fudan.edu.cn}}) and Yinsheng Li.}
\endgroup

\begin{abstract}
    In generalized few-shot 3D point cloud segmentation, support annotations are often compromised by sparsity and occlusion. Blindly adapting to such low-quality samples risks learning spurious geometric patterns rather than robust novel concepts. Although 3D Vision-Language Models have been introduced to supplement supervision, the potential of uncertainty to actively govern adaptation dynamics remains underexplored. Existing signal-level methods like loss reweighting fail to prevent unreliable episodes from steering the optimization trajectory. To address this, we propose \textbf{Uncertainty-Gated Adaptive (UGA)}, to the best of our knowledge the first framework that elevates uncertainty to update-level gating for generalized few-shot 3D point cloud segmentation. The core module, \textbf{Uncertainty-Gated Update (UG-Update)}, estimates episodic reliability via intra-class dispersion over merged support and pseudo-labeled novel supervision, and selectively gates residual gradient updates. This encourages the model to learn novel semantics from reliable episodes while attenuating updates during unreliable ones. Complementing this, we introduce an \textbf{Uncertainty-Based Pseudo-Label Filter (UB-Filter)} to enforce learning from high-confidence dense signals. On ScanNet200 and ScanNet++ (1-/5-shot), UGA sets new state of the art with an average $+3.08$ HM IoU over the strongest prior baseline. Code: \url{https://anonymous.4open.science/r/UGA/}.
\end{abstract}

\section{Introduction}

Semantic segmentation of 3D point clouds serves as a fundamental engine for embodied Artificial Intelligence (AI) and multimedia retrieval. To reduce the prohibitive cost of dense 3D annotation, Generalized Few-Shot 3D Segmentation (GFS-3DS) requires a model to acquire novel classes from only a few support examples while preserving its recognition ability on base classes \cite{An2025}.

Given the inherent sparsity, occlusion, and viewpoint limitations of 3D scanning, a support shot is frequently a fragmented partial view or a noisy outlier rather than a canonical class prototype \cite{jiang2025dynamic}. Blindly forcing the model to adapt to such compromised support samples inevitably results in the learning of spurious geometric features, leading to severe semantic drift and poor generalization \cite{Xu2023,Tsai2024}. The core limitation of existing adaptation pipelines is that they treat every episode as an equally trustworthy optimization event. Although uncertainty has been widely used for signal-level calibration, such as loss reweighting \cite{Kendall2018}, filtering pseudo-labels \cite{Wang2022}, and reliability benchmarking in 3D scenes \cite{Calib3D2025}, these methods still allow every episode to induce an update, implicitly assuming that the resulting optimization direction is safe. We argue that when the global geometry of episode-level novel supervision is highly dispersed or non-representative, signal-level adjustments are insufficient. In such cases, the optimization trajectory is fundamentally misguided. Robust novel-class adaptation therefore requires selective plasticity: the model should release novel-directed updates only when the current episode provides coherent and reliable supervision.

\begin{wrapfigure}[15]{r}[0pt]{0.48\columnwidth}
\begin{minipage}[t]{0.48\columnwidth}
\centering
\includegraphics[width=0.82\linewidth]{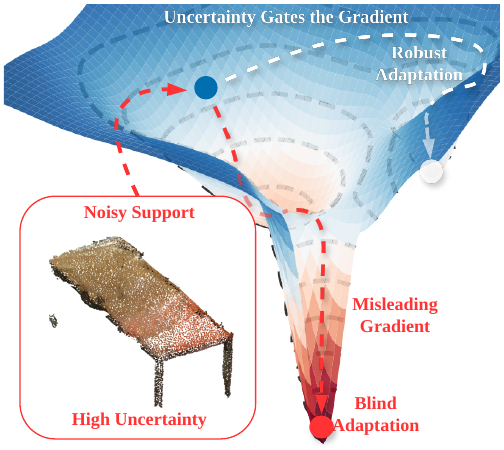}
\vspace{2pt}
\begingroup
\setlength{\abovecaptionskip}{0pt}
\setlength{\belowcaptionskip}{0pt}
\captionsetup{font=small,labelfont={small,bf},textfont=small,justification=centering,singlelinecheck=true,margin=0pt,width=\linewidth}
\caption{\mbox{Blind adaptation vs.\ robust adaptation.}}
\label{fig:intro_ugac}
\endgroup
\end{minipage}
\end{wrapfigure}%
As illustrated in Figure~\ref{fig:intro_ugac}, prior pipelines typically update the model after every support episode, even when the observed geometry is fragmented or non-representative. To address this limitation, we propose \textbf{Uncertainty-Gated Adaptive (UGA)}, a framework that elevates uncertainty from signal-level calibration to update-level control. UGA estimates episodic reliability by measuring the intra-class dispersion of novel features after merging sparse support annotations with pseudo-labeled supervision. It then decomposes the adaptation gradient into a base-aligned component, supported by the Phase-1 optimization geometry, and a residual component that captures novel-directed adaptation. The proposed \textbf{Uncertainty-Gated Update (UG-Update)} module uses episodic uncertainty to regulate this residual component: compact novel supervision preserves residual adaptation, whereas dispersed supervision attenuates unstable residual updates and pulls the update toward the base-aligned direction, reducing the risk of semantic drift. This allows adaptation to remain plastic when supervision is reliable while becoming conservative when the episode is likely to induce misleading gradients.

While UGA prevents the corruption of learned representations by rejecting low-quality updates, this conservatism may reduce the total volume of supervision. To compensate, we leverage 3D Vision-Language Models (VLMs) to generate dense pseudo-supervision for novel regions, a direction that has shown promise in recent open-vocabulary tasks \cite{An2025ICLR,Reason3D2025}. However, consistent with our rigorous quality standards, we recognize that VLMs can also produce hallucinations in cluttered 3D scenes. Therefore, we integrate a complementary \textbf{Uncertainty-Based Pseudo-Label Filter (UB-Filter)}. UB-Filter retains only high-confidence VLM predictions, forming a two-level uncertainty design, UB-Filter purifies the supervision entering the loss, whereas UG-Update regulates the residual gradient induced by that supervision. Our contributions are summarized as follows:
\begin{itemize}
    \item \textbf{Update-level uncertainty gating:} We propose UGA with UG-Update, which uses episodic reliability to regulate residual gradients and prevent noisy adaptation episodes from causing semantic drift. To the best of our knowledge, this is the first work to leverage uncertainty for update-level gating in GFS-3DS.
    \item \textbf{Uncertainty-filtered VLM supervision:} We introduce UB-Filter to retain only reliable VLM pseudo labels, providing high-purity dense signals to mitigate support sparsity.
    \item \textbf{State-of-the-art(SOTA) results:} UGA achieves new SOTA performance on ScanNet200 and ScanNet++ under generalized 1-/5-shot settings~\cite{Rozenberszki2022,ScanNetPP2023}, improving the harmonic mean by an average of +3.08 HM (up to +4.96 HM) over the strongest baseline.
\end{itemize}

\section{Related Work}

\subsection{Generalized Few-Shot 3D Segmentation}
Few-shot segmentation was first extensively studied in 2D images \cite{Shaban2017,Tian2022}, but 3D point clouds present unique challenges due to data sparsity.
Early 3D works adopted episodic meta-learning \cite{Zhao2021} or transductive inference \cite{Ning2023}.
However, real-world deployment requires GFS-3DS, recognizing both base and novel classes \cite{MyersDean2021,An2025}.
Various strategies have been proposed to tackle this, including prototype adaptation \cite{He2023}, decoupling localization from segmentation \cite{Li2024}, and learning ``what not to segment'' to handle backgrounds \cite{Lang2022}.
Others have explored incremental settings \cite{Cermelli2021}, prediction calibration \cite{Lu2023}, and class-contrastive learning \cite{Foo2024} to mitigate bias.
Recent benchmarks have further revisited evaluation protocols \cite{An2024CVPR} and proposed non-parametric networks to reduce training time \cite{Zhu2024}.
While strong fine-tuning baselines \cite{Hajimiri2023}, geometry-aware methods \cite{Xu2023,Tsai2024}, and dynamic routing strategies \cite{jiang2025dynamic} have improved GFS-3DS performance, they largely assume that each adaptation episode provides a trustworthy optimization signal. As a result, episode reliability remains under-modeled, especially when sparse supports or noisy pseudo supervision induce misleading updates.

\subsection{Uncertainty and Reliability in 3D Understanding}
Uncertainty estimation has long been used to improve reliability, from Bayesian approximations such as dropout \cite{Gal2016,Kendall2017} to signal-level calibration in segmentation, including uncertainty-weighted losses \cite{Kendall2018}. In semi-supervised and domain adaptation settings, uncertainty-aware pseudo-label filtering and rectification are standard practices to mitigate noise \cite{Wang2022,Zheng2021,Zou2019,Wang2021}.
In 3D, Calib3D recently established benchmarks for model calibration \cite{Calib3D2025}. However, these approaches predominantly operate at the signal level, modulating input data or loss weights. They do not explicitly control optimization dynamics (e.g., gradients). The proposed UGA framework bridges this gap by elevating uncertainty to the update level, protecting the model from noisy gradient flows.

\subsection{VLM-Assisted 3D Segmentation}
Vision-language models have substantially expanded open-vocabulary 3D understanding by aligning 3D representations with language-supervised visual features \cite{Peng2023,Jiang2024,Rozenberszki2022,Chen2023}. By projecting 3D point clouds to CLIP-aligned feature spaces, these methods enable zero-shot recognition capabilities. In the context of few-shot segmentation, SOTA methods, most notably GFS-VL \cite{An2025} which serves as our primary baseline, leverage VLMs to generate dense pseudo-supervision to compensate for support sparsity \cite{An2025ICLR}. More recently, LLM-driven reasoning has been introduced to refine 3D segmentation results \cite{Reason3D2025}. While promising, VLM-derived supervision can introduce erroneous novel pseudo labels under domain gaps, projection errors, and cluttered 3D observations. Complementing these works, this work introduces UB-Filter to verify VLM signals before using them for adaptation.

\subsection{Safe Adaptation and Gradient Control}
Preventing catastrophic forgetting is a core concern in continual learning. Approaches like Gradient Episodic Memory (GEM) \cite{LopezPaz2017} and its efficient variant A-GEM \cite{Chaudhry2019} constrain updates using memory. Orthogonal Gradient Descent (OGD) \cite{Farajtabar2020} projects gradients to avoid interference in continual learning. More recently, gradient/subspace-based update control has been further explored, such as continual gradient low-rank projection fine-tuning (GORP) \cite{wang-etal-2025-continual}. In 2D segmentation, orthogonality-constrained prototype learning is introduced to reduce base--novel interference \cite{Liu2023}. In contrast, UGA uses episodic uncertainty to regulate the residual update outside the base-aligned subspace. This provides an episode-adaptive form of gradient control, whereas prior gradient-control methods are typically episode-invariant and signal-level uncertainty methods do not regulate the update direction.


\section{Methodology}
\label{sec:method}
\subsection{Overview}
\label{subsec:ugac_overview}
\begin{figure*}[t]
\centering
\includegraphics[width=0.85\textwidth]{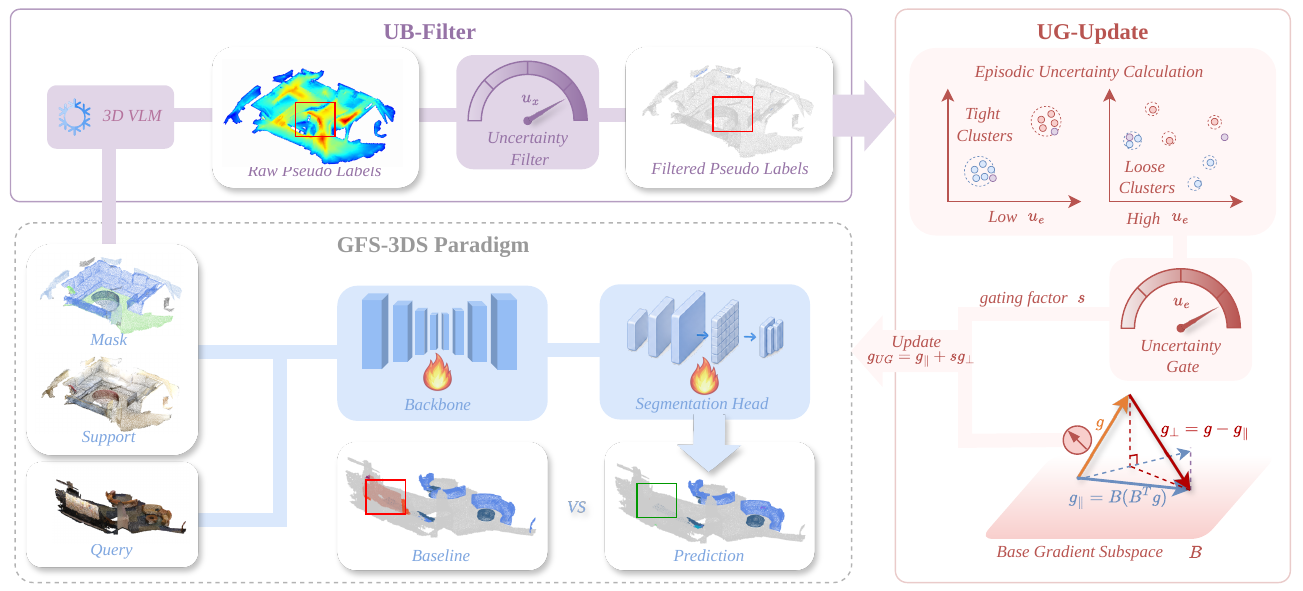}
\caption{Phase~2 UGA schematic: \textbf{UB-Filter} (left) filters dense VLM pseudo labels; \textbf{UG-Update} (right) gates the residual gradient $\mathbf{g}_\perp$. Full description is shown in Sec.~\ref{subsec:ugac_overview}.}
\label{fig:framework}
\end{figure*}

We propose UGA, a unified framework for GFS-3DS that stabilizes novel-class adaptation while preserving base-class semantics under limited supervision. UGA follows a standard two-phase protocol with base pretraining (Phase~1) and novel adaptation (Phase~2), and introduces two complementary uncertainty-aware mechanisms that are activated only during Phase~2. The proposed framework is illustrated in Figure~\ref{fig:framework}. The first component, UG-Update, operates at the optimization level by decomposing the adaptation gradient into a base-aligned component and a residual component, and adaptively suppressing the residual update under high episodic uncertainty. The second component, UB-Filter, operates at the supervision level by retaining only low-uncertainty pseudo-labeled points from a frozen VLM. Together, these two components decouple supervision reliability from update dynamics and effectively mitigate base--novel interference in GFS-3DS. In each Phase~2 batch, UB-Filter first removes high-uncertainty pseudo labels. UG-Update then computes dispersion on the merged support/retained-pseudo supervision and gates the resulting update. The two modules act on orthogonal axes of the same proximal step (see details in Appendix~\ref{app:proximal}). UG-Update is a trust region on the update direction via $\lambda(u_{\mathrm{e}})$, while UB-Filter is a quality filter on the supervision content feeding the gradient $g$ itself. They also use distinct uncertainty sources (intra-class backbone dispersion vs.\ frozen-VLM logits), conflating them loses $2.16$ HM (see details in Sec.~\ref{subsec:ablation}, Table~\ref{tab:ablation_main}).

\subsection{Uncertainty-Gated Update (UG-Update)}
\label{subsec:ug_update}
\noindent\textbf{Motivation.}
Generalized few-shot adaptation is sensitive to unreliable supervision, since even a small fraction of noisy novel signals can steer gradients toward directions that interfere with base semantics. UG-Update addresses this issue by explicitly separating a base-aligned update component (consistent with Phase~1 optimization geometry) from a residual component (plastic but potentially interfering), and then using episodic uncertainty to regulate how much residual update is allowed.

\noindent\textbf{Episodic Uncertainty from Intra-Class Dispersion.}
Let $\mathcal{F}=\{f_i\in\mathbb{R}^{D}\}_{i=1}^{N}$ denote the point-wise embeddings extracted by the segmentation backbone for a batch of $N$ points, where $D$ is the feature dimension. Let $y_i\in\{-1,0,\dots,K-1\}$ be the merged supervision label of point $i$ after combining sparse support annotations with retained VLM pseudo labels, where $-1$ indicates ignored/background or filtered-out points and $K=K_b+K_n$ is the total number of base and novel classes.

We compute episodic uncertainty using only novel-class points from this merged supervision. For each novel class $c\ge K_b$, let $I_c=\{i\mid y_i=c\}$ be its index set ($|I_c|$ counts points with label $c$ in the current batch). We define the eligible novel-class set $\mathcal{C}$ and the per-class prototype $\mu_c$ as
\begin{equation}
\mathcal{C}=\bigl\{c \mid c\ge K_b,\; |I_c|\ge 2 \bigr\},
\qquad
\mu_c=\frac{1}{|I_c|}\sum_{i\in I_c} f_i .
\label{eq:eligible_proto}
\end{equation}
We then measure intra-class dispersion by the average squared Euclidean distance to the prototype. Aggregating across all eligible novel classes yields the raw episodic uncertainty:
\begin{equation}
u_{\mathrm{e,raw}}
= \frac{1}{|\mathcal{C}|}
\sum_{c\in\mathcal{C}}
\frac{1}{|I_c|}
\sum_{i\in I_c}
\|f_i-\mu_c\|_2^2 .
\label{eq:uepi_raw}
\end{equation}
Few-shot learners commonly summarize each class by pooled support embeddings~\cite{Snell2017}, and 3D GFS instantiates the same abstraction via support-feature prototypes~\cite{Zhao2021,An2025}. Thus, Eq.~\eqref{eq:uepi_raw} measures episodic within-cluster scatter around $\mu_c$, corresponding to the average squared residual under a single-centroid summary. Large values indicate that the supervised novel features are poorly represented by one prototype, which violates the compact-geometry premise of prototype-guided adaptation and therefore provides an episode-level reliability signal.
High dispersion can arise from fragmented scans, occluded geometry, or residual pseudo-label errors, all of which make the current novel supervision internally inconsistent and therefore unreliable for steering gradient updates.
The class assignments may come from support annotations or retained pseudo labels, but the scalar itself is computed from backbone feature geometry rather than frozen-VLM confidence.
If $\mathcal{C}=\emptyset$, we set $u_{\mathrm{e}}=0$ for Eq.~\eqref{eq:gug}, so $s\approx 1$ under our default $(k,\tau)$ and the update matches the baseline.

\noindent\textbf{Running normalization.}
To make $u_{\mathrm{e,raw}}$ scale-comparable across episodes, we $z$-normalize it with EMA running statistics $(m_t, v_t)$ of momentum $\beta\in(0,1)$:
\begin{align}
m_t &= \beta m_{t-1} + (1-\beta)\,u_{\mathrm{e,raw}}^{(t)}, \label{eq:rs_mean}\\
v_t &= \beta v_{t-1} + (1-\beta)\big(u_{\mathrm{e,raw}}^{(t)}-m_{t-1}\big)^2, \label{eq:rs_var}\\
u_{\mathrm{e}} &= \big(u_{\mathrm{e,raw}}-m_t\big)\big/\sqrt{v_t+\epsilon}. \label{eq:uepi_norm}
\end{align}

\noindent\textbf{Base Gradient Subspace Construction.}
Let $\phi\in\mathbb{R}^{d}$ denote the vectorized parameter subset $\mathcal{P}_{\mathrm{UG}}$ to be controlled by UG-Update (PTv3 backbone parameters and the base segmentation head; novel-class head excluded), where $d$ is the parameter dimensionality. After Phase~1 training, we extract a representative set of base-task gradients:
\begin{equation}
\mathcal{G}=\{g^{(t)}\in\mathbb{R}^{d}\}_{t=1}^{T},
\label{eq:base_grad_set}
\end{equation}
where $g^{(t)}=\nabla_{\phi}\mathcal{L}_{\mathrm{base}}^{(t)}$ is computed on the $t$-th mini-batch from the base training set and $T$ balances diversity and computational cost. Stacking these gradients yields:
\begin{equation}
G=[g^{(1)},\dots,g^{(T)}]\in\mathbb{R}^{d\times T} .
\end{equation}
We then apply QR factorization:
\begin{equation}
G=QR,\qquad Q^{\top}Q=I ,
\label{eq:basis_qr}
\end{equation}
and retain the first $r$ columns of $Q$ to form an orthonormal basis $B\in\mathbb{R}^{d\times r}$ with $B^{\top}B=I_r$. The basis $B$ remains fixed throughout Phase~2 and spans a base gradient subspace that characterizes Phase~1 update geometry.

\noindent\textbf{Gradient Decomposition and Uncertainty-Gated Residual Update.}
During Phase~2 adaptation, let $g=\nabla_{\phi}\mathcal{L}_{\mathrm{seg}}\in\mathbb{R}^{d}$ denote the gradient induced by the current supervision target. We decompose $g$ into a base-aligned component and an orthogonal residual:
\begin{equation}
g_{\parallel}=B(B^\top g), \qquad
g_{\perp}=g-g_{\parallel}.
\label{eq:decomp}
\end{equation}
Since $B$ is orthonormal, $g_{\parallel}=BB^{\top}g$ is the component of $g$ already supported by the Phase~1 update geometry (i.e., directions the backbone and base classifier have already been driven along during base pretraining), whereas $g_{\perp}$ collects update directions that lie outside this geometry and therefore correspond to \emph{novel-specific} adjustments learned only from Phase~2 supervision. Because GFS-3DS Phase~2 supervision still contains base labels, $g_{\parallel}$ is sourced from supervision the backbone has already been trained on, while $g_{\perp}$ aggregates updates introduced by the few-shot novel supports and VLM pseudo labels, precisely the noisy signals we want to gate when the episode is unreliable. We emphasize that this differs from the continual-learning regime targeted by OGD~\cite{Farajtabar2020}: there, base data is unavailable and the safe direction is $g_{\perp}$ (orthogonal to base, $\langle g^{\text{base}},g_{\perp}\rangle=0$); here, base data is still supervised in Phase~2 and the trustworthy component is $g_{\parallel}$. The two mechanisms are therefore geometric \emph{duals} (see details in Appendix~\ref{app:proximal}).

UG-Update therefore regulates the residual component with an uncertainty-dependent sigmoid gating coefficient $s$, where $k>0$ controls the gating steepness and $\tau$ is the uncertainty threshold (when $u_{\mathrm{e}}>\tau$, $s$ decreases, enforcing a conservative update closer to the base-aligned direction). The controlled gradient $g_{\mathrm{UG}}$ is then
\begin{equation}
s=\sigma\!\left(-k(u_{\mathrm{e}}-\tau)\right),
\qquad
g_{\mathrm{UG}}=g_{\parallel}+s\,g_{\perp}.
\label{eq:gug}
\end{equation}
UG-Update applies Eq.~\eqref{eq:gug} only to a designated parameter subset $\mathcal{P}_{\mathrm{UG}}$, namely the PTv3 backbone parameters and the base segmentation head; the novel-class head is excluded from $\mathcal{P}_{\mathrm{UG}}$ to preserve novel adaptation capacity. Furthermore, Eq.~\eqref{eq:gug} is the closed-form solution of a per-step proximal problem with episode-dependent penalty $\lambda(u_{\mathrm{e}})=e^{k(u_{\mathrm{e}}-\tau)}$ on the orthogonal residual direction (see Appendix~\ref{app:proximal}). The two endpoints of this family are $\lambda{=}0$ (a plain un-projected gradient update on $\mathcal{L}_{\mathrm{seg}}$, instantiated by GFS-VL's standard fine-tuning) and $\lambda\!\to\!\infty$ (a hard projection of the update onto $\mathrm{span}(B)$, the Phase-1 base subspace). The latter is mechanistically \emph{dual} to the orthogonal projection used by OGD~\cite{Farajtabar2020} rather than identical to it (cf.\ Eq.~\eqref{eq:decomp}). We treat OGD as a separate baseline instead of a limit case of UGA (see details in Table~\ref{tab:appendix_ogd_scannet200_1shot}). Rather than committing to either unconstrained adaptation or a hard Phase~1-subspace fallback, the episode-adaptive penalty $\lambda(u_{\mathrm{e}})$ continuously adjusts the update geometry according to episode reliability.

\subsection{Uncertainty-Based Pseudo-Label Filter (UB-Filter)}
\noindent\textbf{Motivation.}
Dense pseudo labels from a frozen VLM can mitigate supervision sparsity in Phase~2, but they are inevitably noisy in cluttered 3D scenes. UB-Filter prevents unreliable pseudo supervision from destabilizing adaptation by retaining only low-uncertainty pseudo-labeled points.

Let the 3D VLM produce per-point logits $\ell(x)\in\mathbb{R}^{K}$ with probabilities
\begin{equation}
p_k(x)=\mathrm{softmax}(\ell(x))_k,\qquad k\in\{0,\dots,K-1\},
\label{eq:vlm_prob}
\end{equation}
where $K=K_b+K_n$. We score point-wise uncertainty $u(x)$ with one of three proxies:
\begin{align}
u_{\mathrm{ent}}(x) &= -\textstyle\sum_{k} p_k(x)\log\!\big(p_k(x)+\epsilon\big), \label{eq:upl_entropy}\\
u_{\mathrm{mar}}(x) &= -\big(p_{(1)}(x)-p_{(2)}(x)\big), \label{eq:upl_margin}\\
u_{\mathrm{max}}(x) &= -\max_{k} p_k(x), \label{eq:upl_maxlogit}
\end{align}
where $p_{(1)},p_{(2)}$ are the top-2 probabilities and $\epsilon>0$ ensures numerical stability. Entropy captures full-distribution dispersion, margin captures top-2 competition, and max-prob captures peak confidence.

Given a retention ratio $\rho\in(0,1)$, UB-Filter performs class-wise quantile filtering over novel pseudo labels. For each novel class $c\in\{K_b,\dots,K-1\}$, we define the set of pseudo-label-$c$ points $\mathcal{X}_c$ and the class-specific threshold $q_c$ as
\begin{equation}
\mathcal{X}_c=\{x \mid \hat{y}(x)=c\},
\qquad
q_c=\mathrm{Q}_{\rho}\big(\{u(x)\mid x\in\mathcal{X}_c\}\big),
\label{eq:quantile}
\end{equation}
where $\hat{y}(x)=\arg\max_k p_k(x)$ is the VLM-predicted pseudo label; we then retain only points satisfying $u(x)\le q_c$, discarding the rest. This guarantees that each novel class contributes a controlled amount of pseudo supervision and avoids bias toward classes with large spatial extent.

\subsection{Training Objective}
\label{subsec:training_objective}
Phase~2 uses the standard point-wise cross-entropy over logits $\mathbf{z}(x)$ (same functional form as the baseline), where $\Omega$ is the mini-batch and $\tilde{y}(x)$ merges sparse supports with VLM pseudo labels elsewhere; UB-Filter maps only uncertain pseudo-labeled points to ignore, while UG-Update replaces $\nabla_{\phi}\mathcal{L}_{\mathrm{seg}}$ by Eq.~\eqref{eq:gug} after backpropagation (Appendix~\ref{sec:algo_impl} gives pseudocode for both):
\begin{equation}
\mathcal{L}_{\mathrm{seg}}
=
\frac{1}{|\Omega|}\sum_{x\in\Omega}\mathrm{CE}\big(\mathbf{z}(x), \tilde{y}(x)\big).
\label{eq:seg_loss}
\end{equation}

\section{Experiments}
\label{sec:exp}

\subsection{Experimental Setup}
\label{sec:expsetup}
We evaluate UGA on ScanNet200~\cite{Rozenberszki2022} and ScanNet++~\cite{ScanNetPP2023} under GFS-3DS~\cite{An2025}. We use disjoint base and novel class splits, as detailed in Appendix~\ref{app:benchmark}. Following the GFS-VL protocol, models are pretrained on base classes, adapted with episodic 1-/5-shot novel supports, and evaluated on query scenes spanning all classes. We report comparisons against the strongest GFS-VL baseline.
We report mIoU (\%) on base (\textbf{B}), novel (\textbf{N}), and all (\textbf{A}) classes, plus harmonic mean (\textbf{HM}) between base and novel mIoU as the stability--plasticity summary.
Unless stated otherwise we follow GFS-VL's protocol using PTv3~\cite{Wu2024PTv3} and frozen RegionPLC~\cite{Yang2024RegionPLC}, with experiments on 4 NVIDIA A100 GPUs averaged over 5 runs; Appendix~\ref{app:seed_stats} reports mean$\pm$std over the five runs.
UGA runs only in Phase~2 ($<1\%$ adaptation-time overhead).
UG-Update applies the sigmoid gate in Eq.~\eqref{eq:gug} with default $(k,\tau)=(7,1.8)$ throughout; sensitivity to $(k,\tau)$ is reported in Appendix~\ref{app:hyper_sens_ugupdate}. UGA-Entropy/Margin/Max differ only in the point-wise uncertainty for UB-Filter (Eqs.~\eqref{eq:upl_entropy}--\eqref{eq:upl_maxlogit}).
Remaining hyperparameters and reproducibility notes are described in Appendix~\ref{app:impl_repro}.

\begin{table*}[t]
\centering
\caption{Results on ScanNet/++ (1/5-shot). Best: \textbf{bold}; second: \underline{underline}.}
\label{tab:main_results}
\resizebox{\textwidth}{!}{%
\begin{tabular}{l|cccc|cccc||cccc|cccc}
\toprule
\multirow{3}{*}{Method}
& \multicolumn{8}{c||}{ScanNet200}
& \multicolumn{8}{c}{ScanNet++} \\
\cmidrule(l{0pt}r{0pt}){2-9}\cmidrule(l{0pt}r{0pt}){10-17}
& \multicolumn{4}{c|}{5-shot} & \multicolumn{4}{c||}{1-shot}
& \multicolumn{4}{c|}{5-shot} & \multicolumn{4}{c}{1-shot} \\
\cmidrule(l{0pt}r{0pt}){2-5}\cmidrule(l{0pt}r{0pt}){6-9}\cmidrule(l{0pt}r{0pt}){10-13}\cmidrule(l{0pt}r{0pt}){14-17}
& B & N & A & HM & B & N & A & HM & B & N & A & HM & B & N & A & HM \\
\midrule
Fully Supervised
& 68.70 & 39.32 & 45.51 & 50.02
& 68.70 & 39.32 & 45.51 & 50.02
& 65.45 & 37.24 & 48.53 & 47.47
& 65.45 & 37.24 & 48.53 & 47.47 \\
\midrule
PIFS~\cite{Cermelli2021}
& 28.78 & 3.82  & 9.07  & 6.71
& 17.84 & 2.87  & 6.02  & 4.88
& 39.98 & 5.74  & 19.44 & 10.03
& 36.66 & 4.95  & 17.63 & 8.71  \\
attMPTI~\cite{Zhao2021}
& 37.13 & 4.99  & 11.76 & 8.79
& 54.84 & 3.28  & 14.14 & 6.17
& 55.89 & 4.19  & 24.87 & 7.78
& 53.16 & 3.55  & 23.40 & 6.66  \\
COSeg~\cite{An2024CVPR}
& 57.67 & 5.21  & 16.25 & 9.54
& 47.03 & 4.03  & 13.09 & 7.42
& 59.34 & 6.96  & 27.91 & 12.45
& 58.49 & 6.24  & 27.14 & 11.26 \\
GW~\cite{Xu2023}
& 59.28 & 8.30  & 19.03 & 14.55
& 55.23 & 6.47  & 16.74 & 11.56
& 51.35 & 11.03 & 27.16 & 18.15
& 46.71 & 6.63  & 22.66 & 11.59 \\
\midrule
\rowcolor{gray!10}
GFS-VL~\cite{An2025}
& 67.57 & 31.67 & 39.23 & 43.12
& 68.48 & 29.18 & 37.45 & 40.92
& 60.05 & 21.66 & 37.02 & 31.82
& 61.39 & 19.42 & 36.21 & 29.47 \\
\rowcolor{gray!10}
GFS-VL (Reproduced)
& 67.30 & 31.52 & 39.05 & 42.92
& 67.64 & 28.80 & 36.98 & 40.39
& 61.24 & 21.63 & 37.47 & 31.93
& 60.75 & 18.61 & 35.47 & 28.47 \\
\midrule
\rowcolor{gray!10}
UGA-Margin
& \third{68.29} & \third{32.03} & \third{39.67} & \third{43.61}
& \second{69.14} & \third{29.25} & \third{37.65} & \third{41.11}
& \second{63.80} & \second{24.20} & \second{40.04} & \second{35.07}
& \second{62.27} & \second{20.21} & \second{36.62} & \second{30.50} \\
\rowcolor{gray!10}
UGA-Max
& \second{68.44} & \second{32.41} & \second{39.99} & \second{43.98}
& \best{69.30} & \second{30.16} & \second{38.40} & \second{42.02}
& \best{64.78} & \third{23.55} & \second{40.04} & \third{34.53}
& \best{63.26} & \third{19.76} & \third{35.16} & \third{30.08} \\
\rowcolor{gray!10}
UGA-Entropy
& \best{69.36} & \best{34.02} & \best{41.46} & \best{45.65}
& \third{68.89} & \best{32.14} & \best{39.88} & \best{43.83}
& \third{63.43} & \best{25.92} & \best{40.93} & \best{36.78}
& \third{61.73} & \best{21.06} & \best{37.33} & \best{31.39} \\
\bottomrule
\end{tabular}%
}
\end{table*}

\subsection{Comparison with SOTAs}
\label{subsec:sota}
\noindent\textbf{Quantitative results.}
Table~\ref{tab:main_results} compares the Fully Supervised oracle (upper bound; from GFS-VL~\cite{An2025}), prior few-shot methods (PIFS, attMPTI, COSeg, GW), GFS-VL, our local re-run of GFS-VL, and UGA on ScanNet200 / ScanNet++.
UGA-Entropy wins every split, lifting HM by $+2.53$ / $+2.91$ / $+4.96$ / $+1.92$ over GFS-VL (avg.\ $+3.08$), and the gains concentrate on Novel mIoU ($+2.35$ / $+2.96$ / $+4.26$ / $+1.64$), showing that our gating actively unlocks novel-class capacity rather than merely guarding the base.
Relative to the Fully Supervised oracle, UGA reaches $91\%$ / $88\%$ of oracle HM on ScanNet200 (5-/1-shot) and $77\%$ / $66\%$ on ScanNet++, substantially narrowing the few-shot-to-full-annotation gap.
Classical few-shot baselines (PIFS/attMPTI/COSeg/GW) rely only on limited few-shot samples and cap out at HM~$\le 19$, confirming the need for VLM-assisted dense supervision in GFS-3DS.
Gains also transfer to the standard ScanNet and to SCN/OpenScene backbone/VLM swaps (as shown in Appendices~\ref{app:scannet_split},~\ref{app:generality_tables}), confirming a structural, dataset/model-agnostic effect, with a direct OGD comparison in Appendix~\ref{app:proximal} ($+2.80$ HM over OGD).

\begin{figure}[t]
  \centering
  \includegraphics[width=\linewidth]{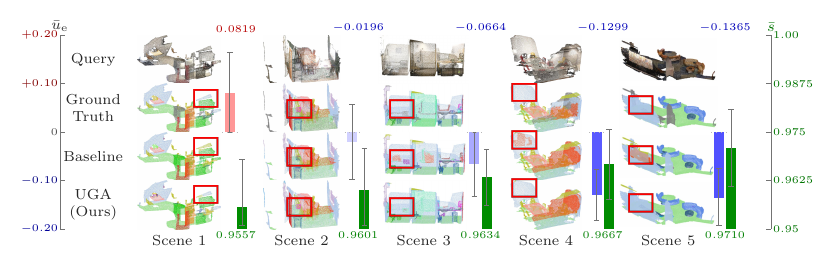}
  \caption{Per-scene qualitative results on $5$ ScanNet200 episodes, $\bar u_{\mathrm{e}}$ (left bar) and gate $\bar s$ (right bar; whiskers $\pm 1\sigma$ over 5 runs). Red boxes highlight regions where UGA suppresses mutual base--novel cross-contamination.}
  \label{fig:main_qual}
  \vspace{-8pt}
\end{figure}

\noindent\textbf{Qualitative results.}
As $\bar u_{\mathrm{e}}$ decreases left-to-right, $\bar s$ opens monotonically. Its absolute level reflects the per-step variability of $u_{\mathrm{e}}$ on noisy mini-batches (Jensen: $\bar s\!\neq\!\sigma(-k(\bar u_{\mathrm{e}}-\tau))$ since $\sigma$ is nonlinear), consistent with the per-run $\bar s\!\in\![0.94,0.98]$ measured across our experiments (per-step response is shown in Fig.~\ref{fig:ugupdate_grad_analysis}(b)). UGA adapts its gate across scenes and reduces base--novel cross-contamination in the red-boxed regions relative to GFS-VL, consistent with Table~\ref{tab:main_results}.

\begin{table}[t]
\centering
\caption{Module ablation on ScanNet200 5-shot, mIoU (\%). \textbf{(a)} Components on/off (UG-Update gate proxy = our default intra-class dispersion). \textbf{(b)} Both components active; varying the UG-Update gate proxy. \textbf{Bold}: best per column; \underline{underline}: second; plain: third or below.}
\label{tab:ablation_main}
\setlength{\tabcolsep}{5pt}
\renewcommand{\arraystretch}{1.06}
\small
\begin{tabular}{cc|l|cccc}
\toprule
UG-Update & UB-Filter & Gate proxy & B & N & A & HM \\
\midrule
\multicolumn{7}{l}{\textit{(a) Components on/off}} \\
            &            & --                              & 67.57 & 31.67 & 39.23 & 43.12 \\
\checkmark  &            & dispersion                      & 68.06 & 32.00 & 39.59 & 43.53 \\
            & \checkmark & --                              & \third{68.81} & \third{32.34} & \third{40.02} & \third{44.00} \\
\midrule
\multicolumn{7}{l}{\textit{(b) Both components active; varying UG-Update gate proxy}} \\
\checkmark  & \checkmark & VLM Agg.\ Entropy               & 68.42 & 31.87 & 39.57 & 43.49 \\
\checkmark  & \checkmark & VLM Agg.\ Margin                & \second{69.00} & 32.27 & 40.00 & 43.97 \\
\checkmark  & \checkmark & VLM Agg.\ Max-Prob              & 68.50 & \second{32.87} & \second{40.37} & \second{44.42} \\
\checkmark  & \checkmark & Intra-class dispersion (ours)   & \best{69.36} & \best{34.02} & \best{41.46} & \best{45.65} \\
\bottomrule
\end{tabular}
\end{table}

\subsection{Ablation Studies}
\label{subsec:ablation}

Table~\ref{tab:ablation_main} reports the ablation study on the ScanNet200 5-shot setting, with consistent trends observed across the other three splits. The complete multi-split results will be included in the public release. \textbf{(a)~Components on/off.} UG-Update and UB-Filter each provide independent gains, improving HM by $+0.41$ and $+0.88$ points, respectively. When combined, they achieve $45.65$ HM, improving the baseline by $+2.53$ points and exceeding the sum of their standalone gains by $+1.24$ points, confirming that cleaner supervision (UB-Filter) and reliable update geometry (UG-Update) act on orthogonal axes of the same proximal step (see details in Sec.~\ref{subsec:ugac_overview}). \textbf{(b)~Gate proxy.} When both modules are used, the appropriate reference is the UB-Filter-only variant ($44.00$ HM), rather than the original GFS-VL baseline ($43.12$ HM). Relative to this reference, VLM-derived aggregate uncertainty proxies provide limited or inconsistent gains, with Agg.\ Entropy, Agg.\ Margin, and Agg.\ Max-Prob changing HM by $-0.51$, $-0.03$, and $+0.42$ points, respectively. In contrast, the proposed intra-class dispersion proxy improves HM by $+1.65$ points, supporting our design choice to decouple the UG-Update gating signal from the frozen-VLM confidence used by UB-Filter. This separation is important because VLM confidence is largely fixed during Phase~2, whereas feature dispersion reflects the evolving backbone representation and is better aligned with prototype-guided adaptation.


\subsection{UG-Update Mechanism Analysis}
\label{subsec:ugupdate_mech}

\begin{figure}[t]
  \centering
  \includegraphics[width=0.8\columnwidth]{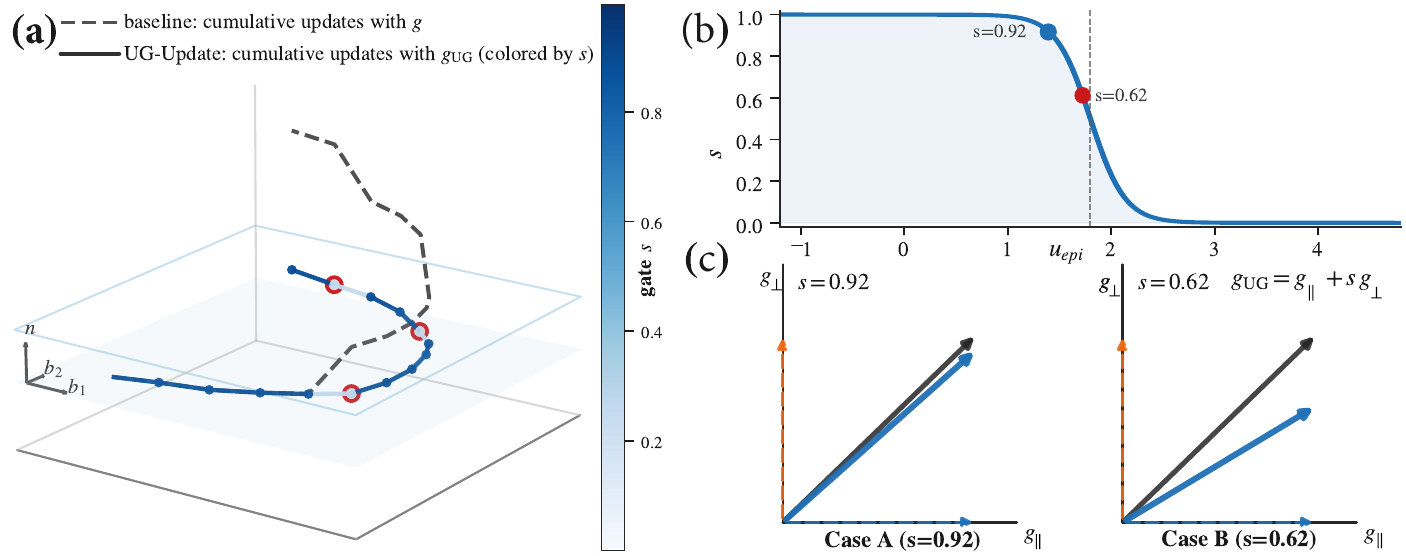}
  \caption{UG-Update: (a) gradient visualization; (b) gate $s$ vs.\ $u_{\mathrm{e}}$; (c) gated decomposition.}
  \vspace{-10pt}
  \label{fig:ugupdate_grad_analysis}
\end{figure}

Fig.~\ref{fig:ugupdate_grad_analysis} illustrates how UG-Update translates episodic uncertainty into controlled gradient updates over the course of adaptation. \textbf{(b)} The sigmoid gate $s$ stays near $1$ on reliable (compact-feature) episodes and drops toward $0$ as dispersion rises. \textbf{(c)} A reliable episode ($s{\approx}0.92$) keeps $g_{\mathrm{UG}}$ close to the full $g$, maximizing plasticity; an unreliable one ($s{\approx}0.62$) shrinks $g_{\perp}$, pulling the update toward the Phase-1 subspace and reducing base interference. \textbf{(a)} This attenuation keeps the optimization trajectory aligned with the base directions $(b_1, b_2)$ while limiting drift along the residual axis $n$. This behavior visualizes an adaptive tightening of the proximal trust region $M=I+\lambda(u_{\mathrm{e}})P_{\perp}$. The resulting update lies between plain adaptation with $\lambda=0$ and a hard Phase~1-subspace projection as $\lambda\to\infty$, which is the geometric dual of OGD. Appendix~\ref{app:proximal} compares against true OGD as a separate baseline, and Appendix~\ref{app:hyper_sens_ugupdate} reports sensitivity to $(k,\tau)$.

\subsection{UB-Filter Mechanism Analysis}
\label{subsec:ubfilter_mech}

\begin{figure*}[t]
  \centering
  \includegraphics[width=0.8\textwidth]{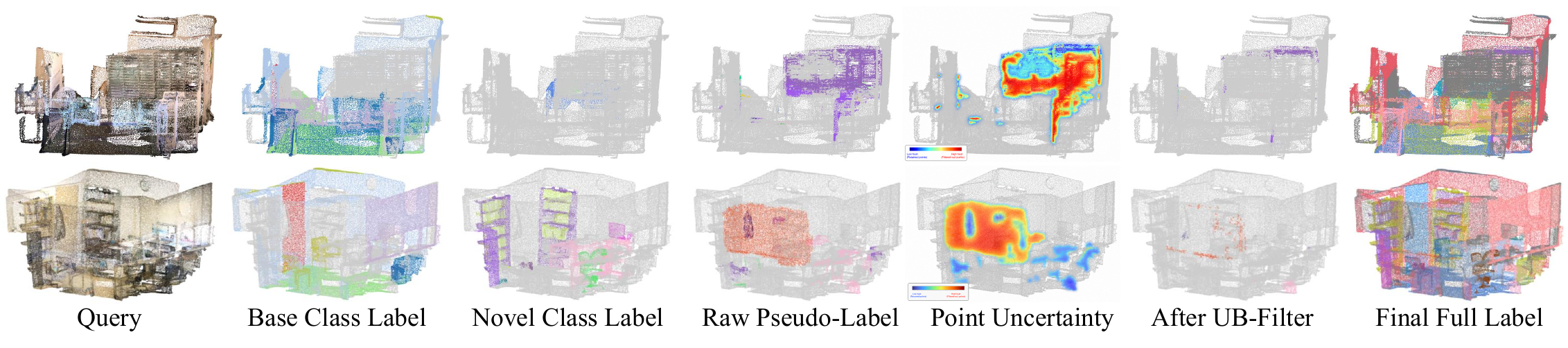}
  \caption{
Visualization of pseudo-label refinement via UB-Filter. Each row shows a query scene.
}
  \label{fig:ubfilter_vis}
\end{figure*}

\begin{figure}[t]
  \centering
  \includegraphics[width=0.8\columnwidth]{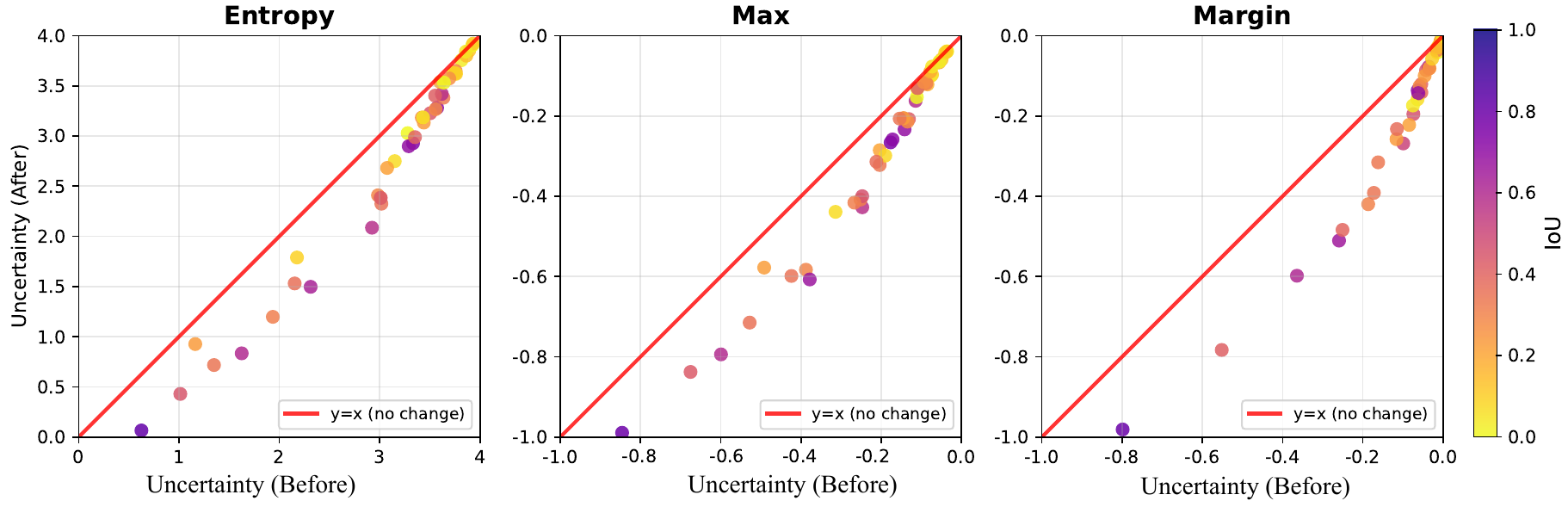}
  
  \caption{UB-Filter. Class-mean uncertainty before ($x$) vs.\ after ($y$); color=IoU; red $y{=}x$.}
  \vspace{-8pt}
  \label{fig:fig_unpl_before_after_1x3}
\end{figure}

Fig.~\ref{fig:ubfilter_vis} compares raw and filtered pseudo supervision before it is merged with support annotations, with each row corresponding to one scene. UB-Filter ranks novel pseudo-label points using entropy, margin, or max-logit scores (Eqs.~\eqref{eq:upl_entropy}--\eqref{eq:upl_maxlogit}) and removes points that are not representative of reliable novel-class supervision.
Fig.~\ref{fig:fig_unpl_before_after_1x3} summarizes class-wise mean uncertainty before versus after filtering for each proxy.
Tables~\ref{tab:main_results}--\ref{tab:ablation_main} quantify performance. Table~\ref{tab:ablation_main} isolates UB-Filter with HM gains (e.g., +0.88 on ScanNet200 5-shot), consistent with suppressing scattered low-confidence regions while retaining compact confident novel areas.

\noindent\textbf{Precision-oriented design against false positives.}
In GFS-3DS, false-positive pseudo labels (base/background points wrongly tagged as novel) are far more harmful than false negatives (missed novel points covered by real supports), because they inject spurious semantic gradients into the novel head. UB-Filter is designed to reduce false-positive pseudo labels. For each novel class, it retains only the $\rho$ fraction of points with the lowest uncertainty scores according to Eq.~\eqref{eq:quantile} and maps the remaining points to \texttt{ignore}. This procedure removes pseudo labels from high-uncertainty regions, where VLM errors are more likely to occur. Fig.~\ref{fig:fig_unpl_before_after_1x3} supports this interpretation by showing reduced mean uncertainty and improved post-filter IoU for classes with cluttered base/background predictions, while confident novel-class cores remain largely unchanged. This pattern suggests that UB-Filter primarily removes the high-uncertainty tail rather than uniformly thinning pseudo labels.

\noindent\textbf{Uncertainty proxies.}\label{subsubsec:ubfilter_before_after}
Fig.~\ref{fig:fig_unpl_before_after_1x3} aligns each proxy with Table~\ref{tab:main_results}.
\textbf{Entropy} (Eq.~\eqref{eq:upl_entropy}) places points mostly below the $y{=}x$ diagonal with darker IoU toward low uncertainty, and attains the best HM on all four splits (+2.53 / +2.91 / +4.96 / +1.92 vs.\ GFS-VL), matching our default.
\textbf{Margin} (Eq.~\eqref{eq:upl_margin}) shows the same diagonal trend but delivers the smallest HM gains among proxies (+0.19--3.25), because top-2 separation is a coarse boundary cue under a fixed $\rho$.
\textbf{Max logit} (Eq.~\eqref{eq:upl_maxlogit}) gains +0.61--2.71 HM and trails entropy on ScanNet++. Its Base mIoU is highest on ScanNet200 1-shot and both ScanNet++ settings ($69.30$, $64.78$, and $63.26$), indicating that peak-confidence filtering favors base-dominated regions while entropy pools ambiguity more effectively for HM.

\section{Conclusion}

We propose UGA, the first GFS-3DS framework to actively regulate adaptation dynamics using update-level uncertainty for stable few-shot 3D adaptation under noisy, sparse supervision. UGA combines UG-Update, which gates residual gradients by episodic reliability, and UB-Filter, which retains reliable VLM pseudo-labels via class-wise uncertainty filtering. Achieving SOTA performance on ScanNet200 and ScanNet++ (1-/5-shot), UGA demonstrates a superior stability–plasticity trade-off. Beyond GFS-3DS, using uncertainty to reshape optimization geometry, rather than simply reweighting signals, holds broad relevance for continual learning and other sparse-supervision regimes. \textbf{Limitations.} Validated on indoor RGB-D; outdoor LiDAR requires re-estimating the base-gradient subspace, and UB-Filter cannot correct confidently wrong pseudo-labels (affecting both loss and dispersion estimate). \textbf{Reproducibility / broader impact.} Implementation, splits, pseudocode, and compute budget are in Appendices~\ref{app:impl_repro},~\ref{app:benchmark},~\ref{sec:algo_impl}; UGA targets indoor 3D scene understanding (robotics, AR/VR, accessibility) with risks inherited from the underlying datasets and pretrained VLMs (class bias, indoor-only validation).

\clearpage
\bibliographystyle{plainnat}  
\small
\bibliography{my_ref}
\normalsize

\clearpage


\appendix

\clearpage
\begin{center}
{\fontsize{28}{34}\selectfont\bfseries Appendix}%
\end{center}
\vspace{1em}
\begingroup
\small
\setlength{\parskip}{0.45em}
\setlength{\parindent}{0pt}
\makeatletter
\def\appendixtocline#1{%
  \noindent\textbf{\ref{#1}}\quad
  \hyperref[#1]{\nameref{#1}}\dotfill\nobreak
  \makebox[1.5em][r]{\pageref{#1}}\par
}%
\makeatother
\appendixtocline{app:proximal}
\appendixtocline{sec:algo_impl}
\appendixtocline{app:benchmark}
\appendixtocline{app:impl_repro}
\appendixtocline{app:scannet_split}
\appendixtocline{app:generality_tables}
\appendixtocline{app:hyper_sens_ugupdate}
\appendixtocline{app:more_vis}
\endgroup
\clearpage

\section{Proximal Interpretation, Limit Cases, and Empirical Validation}
\label{app:proximal}
\label{app:ogd}

\noindent\textbf{Per-step proximal problem.}
The update rule~\eqref{eq:gug} admits a proximal interpretation. Let $P_{\parallel}=BB^\top$ be the orthogonal projector onto $\mathrm{span}(B)$ and $P_{\perp}=I-P_{\parallel}$. For episode-dependent regularization strength $\lambda(u_{\mathrm{e}})\ge 0$, consider
\begin{equation}
\phi_{t+1}
=
\arg\min_{\phi\in\mathbb{R}^{d}}\;
\langle g,\phi-\phi_t\rangle
+\frac{1}{2\eta}\|\phi-\phi_t\|_2^2
+\frac{\lambda(u_{\mathrm{e}})}{2\eta}\|P_{\perp}(\phi-\phi_t)\|_2^2 ,
\label{eq:proximal}
\end{equation}
where $\eta>0$ is the learning rate. The first two terms are a standard linearized proximal step; the third imposes an additional quadratic trust region on the component of the step that lies outside $\mathrm{span}(B)$.

\noindent\textbf{Closed form.}
Setting the gradient of the objective in~\eqref{eq:proximal} to zero yields
\begin{equation}
\phi_{t+1}-\phi_t = -\eta\big[I+\lambda(u_{\mathrm{e}})\,P_{\perp}\big]^{-1}g .
\label{eq:proximal_solve}
\end{equation}
Since $P_{\perp}$ is an orthogonal projector, the matrix $I+\lambda P_{\perp}$ acts as the identity on $\mathrm{span}(B)$ and as $(1+\lambda)\,I$ on its orthogonal complement; its inverse therefore acts as $I$ on $\mathrm{span}(B)$ and as $(1+\lambda)^{-1}I$ off it. Decomposing $g=g_{\parallel}+g_{\perp}$ gives
\begin{equation}
\phi_{t+1}-\phi_t = -\eta\big(g_{\parallel}+s\,g_{\perp}\big),\qquad
s=\big(1+\lambda(u_{\mathrm{e}})\big)^{-1},
\label{eq:proximal_closed}
\end{equation}
which exactly matches~\eqref{eq:gug}. Identifying $s$ with the sigmoid gate component of~\eqref{eq:gug} via $\sigma(-k(u_{\mathrm{e}}-\tau))=(1+\exp(k(u_{\mathrm{e}}-\tau)))^{-1}$ fixes the explicit penalty
\begin{equation}
\lambda(u_{\mathrm{e}})=\exp\!\big(k(u_{\mathrm{e}}-\tau)\big).
\label{eq:lambda_penalty}
\end{equation}

\noindent\textbf{Limit cases.}
The penalty $\lambda(u_{\mathrm{e}})$ governs a continuous spectrum of update geometries:
\begin{itemize}[leftmargin=1.5em,topsep=2pt,itemsep=1pt,parsep=0pt,partopsep=0pt]
\item $\lambda=0$: the residual term in~\eqref{eq:proximal} vanishes, $s=1$, and~\eqref{eq:proximal_closed} reduces to a plain (un-projected) gradient update on $\mathcal{L}_{\mathrm{seg}}$ (instantiated in our pipeline by GFS-VL's standard fine-tuning).
\item $\lambda\to\infty$: the residual term enforces a hard constraint $P_{\perp}(\phi_{t+1}-\phi_t)=0$, $s\to 0$, and~\eqref{eq:proximal_closed} reduces to $-\eta\,g_{\parallel}$, i.e., a hard projection of the update onto the Phase-1 base subspace $\mathrm{span}(B)$.
\item Finite, episode-adaptive $\lambda(u_{\mathrm{e}})$ (UGA): smoothly interpolates between the two limits, dialling the trust region tighter on noisy episodes (large $u_{\mathrm{e}}$) and looser on reliable ones.
\end{itemize}
The corresponding parameter-space preconditioner $M=I+\lambda(u_{\mathrm{e}})P_{\perp}$ matches a Mahalanobis trust geometry whose anisotropy grows along the residual axis with episodic uncertainty (cf.\ Fig.~\ref{fig:ugupdate_grad_analysis}).

\noindent\textbf{Relation to OGD: dual, not generalization.}
The $\lambda\!\to\!\infty$ endpoint above retains $g_{\parallel}$ and discards $g_{\perp}$, which is mechanistically the geometric \emph{dual} of the orthogonal projection used by OGD~\cite{Farajtabar2020} rather than identical to it: OGD prevents catastrophic forgetting by projecting updates onto $g_{\perp}$ (so that $\langle g^{\text{base}},\Delta\phi\rangle=0$, enforcing first-order invariance of the base loss in a continual-learning regime where base data is unavailable), whereas UGA's $\lambda\!\to\!\infty$ limit retains $g_{\parallel}$ because GFS-3DS Phase~2 supervision \emph{still includes base labels}: the segmentation gradient $g$ then naturally decomposes into a Phase-1-consistent component ($g_{\parallel}$, sourced from supervision the backbone has already been trained on) and a novel-specific component ($g_{\perp}$, sourced solely from few-shot supports and VLM pseudo labels). When the episode is unreliable, the principled fallback is to retain $g_{\parallel}$ (which agrees with already-trusted Phase-1 geometry over base supervision) and attenuate $g_{\perp}$ (the entry point for noisy novel supports and VLM hallucinations). OGD's strategy of zeroing $g_{\parallel}$ would discard this base-helpful supervision, which is undesirable in GFS-3DS. UGA's adaptive interior solution should therefore not be read as a soft OGD; we report empirical comparisons against a faithful OGD baseline (i.e., $\Delta\phi=-\eta g_{\perp}$, identical to the original recipe of~\cite{Farajtabar2020}) in Table~\ref{tab:appendix_ogd_scannet200_1shot} as a separate baseline.

\noindent\textbf{Equivalence to gradient-level Tikhonov.}
Replacing the linearized objective in~\eqref{eq:proximal} with the gradient-shaping form $\min_{v\in\mathbb{R}^{d}}\tfrac{1}{2}\|v-g\|_2^2+\tfrac{\lambda(u_{\mathrm{e}})}{2}\|P_{\perp}v\|_2^2$ yields the same minimizer $g_{\mathrm{UG}}$ with the same $s=(1+\lambda)^{-1}$. The two views are therefore equivalent up to the learning-rate factor; we keep~\eqref{eq:proximal} as the primary statement because its limits coincide with established gradient-projection methods.

\noindent\textbf{Empirical comparison against OGD.}
We compare UGA against a faithful implementation of OGD~\cite{Farajtabar2020} on ScanNet200 1-shot. The OGD baseline applies $\Delta\phi=-\eta\,g_{\perp}$ throughout Phase~2; its projection direction $g_{\perp}$ is computed against an orthonormal basis built from Phase~1 logit gradients $\nabla_{\phi}f_{k^{*}}(x;\phi^{*}_{\mathrm{base}})$ as prescribed by the canonical OGD construction~\cite{Farajtabar2020} (which differs from UGA's loss-gradient basis $B$ in Eq.~\eqref{eq:basis_qr}; see Sec.~\ref{subsec:ug_update}). Two predictions follow from the proximal family above. (i)~The plain ($\lambda{=}0$) endpoint, instantiated by GFS-VL's standard fine-tuning, sacrifices stability. (ii)~The Phase-1-subspace projection ($\lambda{\to}\infty$) sacrifices plasticity. Because OGD is the geometric dual of (ii) rather than a member of our family, it occupies a third operating point that is also dominated by an adaptive interior $\lambda(u_{\mathrm{e}})$. Results in Table~\ref{tab:appendix_ogd_scannet200_1shot} are averaged over 5 runs.

\begin{table}[h]
\centering
\caption{Empirical comparison of the plain endpoint ($\lambda{=}0$, GFS-VL) and a faithful OGD baseline (geometric dual of UGA's $\lambda\!\to\!\infty$ limit; not a member of our proximal family) against UGA's finite, episode-adaptive $\lambda(u_{\mathrm{e}})$ on ScanNet200 (1-shot). Results in mIoU (\%). Best: \textbf{bold}; second: \underline{underline}.}
\label{tab:appendix_ogd_scannet200_1shot}
\setlength{\tabcolsep}{6pt}
\renewcommand{\arraystretch}{1.05}
\begin{tabular}{lcccc}
\toprule
Method & B & N & A & HM \\
\midrule
GFS-VL (Reproduced) ($\lambda{=}0$ endpoint) & 67.64 & 28.80 & 36.98 & 40.39 \\
OGD~\cite{Farajtabar2020} ($\Delta\phi=-\eta g_{\perp}$; dual of $\lambda\!\to\!\infty$) & 68.84 & 29.22 & 37.56 & 41.03 \\
\midrule
UGA-Margin (ours)
& \second{69.14}
& \third{29.25}
& \third{37.65}
& \third{41.11} \\
UGA-Max (ours)
& \best{69.30}
& \second{30.16}
& \second{38.40}
& \second{42.02} \\
UGA-Entropy (ours)
& \third{68.89}
& \best{32.14}
& \best{39.88}
& \best{43.83} \\
\bottomrule
\end{tabular}
\end{table}

The plain endpoint ($\lambda{=}0$, instantiated as GFS-VL Reproduced) reaches $40.39$ HM, and the OGD baseline reaches $41.03$ HM---OGD's hard projection improves Base ($+1.20$) with smaller gains on Novel ($+0.42$), consistent with replacing the full update by its orthogonal component throughout adaptation regardless of episode quality. Every UGA variant strictly dominates both points in HM ($41.11$/$42.02$/$43.83$ vs.\ $40.39$/$41.03$), and the best variant (UGA-Entropy) lifts HM by $+2.80$ over OGD with $+2.92$ Novel mIoU---improvements that come from precisely the regime neither rigid choice can represent: a finite, episode-adaptive $\lambda(u_{\mathrm{e}})$ that preserves plasticity on reliable episodes while tightening the trust region on noisy ones. For full disclosure, the OGD row in Table~\ref{tab:appendix_ogd_scannet200_1shot} restricts the orthogonal projection to $\{\texttt{base\_proto},\texttt{classifier}\}$ as in our reproduction of OGD on top of the GFS-VL prototype-based pipeline (OGD~\cite{Farajtabar2020} specifies the gradient construction but not the parameter scope), whereas UGA controls a wider set of parameters (PTv3 backbone $+$ base segmentation head, excluding the novel head); a hypothetical OGD with the same wider $\mathcal{P}$ would still attain $\Delta\phi\!\propto\!g_{\perp}$ on every episode and therefore would still be a non-adaptive endpoint that the interior solution dominates by the same proximal-family argument.

\section{Algorithms and Pseudocode}
\label{sec:algo_impl}

This section summarizes the core procedures used in Phase~2 adaptation. We provide
pseudocode for UG-Update, which performs uncertainty-gated residual gradient control
on a designated parameter subset, and for UB-Filter, which filters dense pseudo labels
by class-wise quantile thresholds computed on-the-fly. Both routines follow the definitions
in Sec.~\ref{sec:method} and use the referenced equations for uncertainty computation, gradient decomposition,
and quantile selection.
They are formalized as step-by-step procedures in Table~\ref{alg:ugupdate} (UG-Update) and Table~\ref{alg:ubfilter} (UB-Filter).

\begin{table}[!htbp]
\centering
\caption{Algorithm: Uncertainty-Gated Residual Update (\textbf{UG-Update}, Phase~2).}
\label{alg:ugupdate}
\setlength{\tabcolsep}{6pt}
\renewcommand{\arraystretch}{1.18}
\small
\begin{tabular}{p{0.98\linewidth}}
\toprule
\textbf{Input:}
Controlled parameters $\phi$ (subset $\mathcal{P}_{\mathrm{UG}}$),
fixed basis $B$,
batch embeddings $\{f_i\}$ and labels $\{y_i\}$,
gate slope $k$,
threshold $\tau$. \\
\textbf{Output:}
Controlled gradient $g_{\mathrm{UG}}$ applied to $\mathcal{P}_{\mathrm{UG}}$. \\
\midrule
\textbf{1. Eligible novel classes.}
Construct $\mathcal{C}$ as in Eq.~\eqref{eq:eligible_proto}. \\[1pt]

\textbf{2. Prototypes.}
For each $c\in\mathcal{C}$, compute
$\mu_c$ using Eq.~\eqref{eq:eligible_proto}. \\[1pt]

\textbf{3. Episodic uncertainty.}
Compute raw uncertainty $u_{\mathrm{e,raw}}$ by feature dispersion
(Eq.~\eqref{eq:uepi_raw}); if enabled, normalize to $u_{\mathrm{e}}$
via Eq.~\eqref{eq:uepi_norm}. \\[1pt]

\textbf{4. Backprop gradient.}
Compute
$g \leftarrow \nabla_{\phi}\mathcal{L}_{\mathrm{seg}}$. \\[1pt]

\textbf{5. Decompose gradient.}
Project onto the base-gradient subspace (Eq.~\eqref{eq:decomp}):
\[
\qquad g_{\parallel} \leftarrow B(B^\top g), \qquad
g_{\perp} \leftarrow g - g_{\parallel}.
\] \\[-2pt]

\textbf{6. Gate coefficient and controlled gradient.}
Compute the gate $s \leftarrow \sigma\!\big(-k(u_{\mathrm{e}}-\tau)\big)$ and the controlled gradient $g_{\mathrm{UG}} \leftarrow g_{\parallel} + s\,g_{\perp}$
(Eq.~\eqref{eq:gug}). \\[1pt]

\textbf{7. Apply to parameter subset.}
Replace the gradient for $\phi\in\mathcal{P}_{\mathrm{UG}}$ by $g_{\mathrm{UG}}$;
keep novel-specific parameters unconstrained. \\
\bottomrule
\end{tabular}
\end{table}

\begin{table}[!htbp]
\centering
\caption{Algorithm: Class-wise Quantile Filtering (\textbf{UB-Filter}, Phase~2).}
\label{alg:ubfilter}
\setlength{\tabcolsep}{6pt}
\renewcommand{\arraystretch}{1.18}
\small
\begin{tabular}{@{}p{0.96\linewidth}@{}}
\toprule
\textbf{Input:}
Per-point VLM probabilities $\{p(x)\}$ (Eq.~\eqref{eq:vlm_prob}),
novel class set $\mathcal{Y}_{\mathrm{novel}}$,
retention ratio $\rho$,
uncertainty function $u(\cdot)$. \\
\textbf{Output:}
Filtered pseudo label map $\tilde{y}_{\mathrm{vlm}}(\cdot)$
(uncertain points mapped to \texttt{ignore}). \\
\midrule
\textbf{1. Pseudo labeling.}
For each point $x$, set
$\hat{y}(x)\leftarrow\arg\max_k p_k(x)$. \\[1pt]

\textbf{2. Uncertainty scoring.}
For each point $x$, compute $u(x)$ using one choice among:
entropy (Eq.~\eqref{eq:upl_entropy}),
margin (Eq.~\eqref{eq:upl_margin}),
or max-probability (Eq.~\eqref{eq:upl_maxlogit}). \\[1pt]

\textbf{3. Initialize output.}
Set $\tilde{y}_{\mathrm{vlm}}(x)\leftarrow \hat{y}(x)$ for all points $x$. \\[1pt]

\textbf{4. Class-wise thresholds (novel only).}
For each $c\in\mathcal{Y}_{\mathrm{novel}}$:
\[
\qquad \mathcal{X}_c \leftarrow \{x\mid \hat{y}(x)=c\},
\qquad q_c \leftarrow \mathrm{Q}_{\rho}\big(\{u(x)\mid x\in\mathcal{X}_c\}\big)
\quad (\text{Eq.~\eqref{eq:quantile}}).
\] \\[-2pt]

\textbf{5. Filtering.}
For each $x\in\mathcal{X}_c$:
if $u(x)>q_c$, set
$\tilde{y}_{\mathrm{vlm}}(x)\leftarrow \texttt{ignore}$. \\
\bottomrule
\end{tabular}
\end{table}

\section{Benchmark Details and Class Splits}
\label{app:benchmark}

We evaluate UGA under the Generalized Few-Shot 3D Segmentation (GFS-3DS) setting on two challenging benchmarks: ScanNet200~\cite{Rozenberszki2022} and ScanNet++~\cite{ScanNetPP2023}. To ensure strict comparability with prior work, we follow the exact protocol introduced in GFS-VL~\cite{An2025}, including class splits and frequency-based category filtering.

\textbf{Category selection protocol.}
Following GFS-VL, categories are selected based on instance occurrence frequency. For ScanNet200, we retain 57 classes that appear more than 100 times. The top 12 most frequent classes are assigned as base, and the remaining 45 as novel. For ScanNet++, 30 classes with more than 80 occurrences are selected, with the top 12 as base and the remaining 18 as novel. This selection ensures adequate supervision for base classes while preserving novel-class diversity and scene-level sparsity.

\textbf{Full class lists.}
The exact class splits used in our experiments are provided below. These are directly inherited from the public GFS-VL repository to ensure consistency.

\textit{ScanNet200}: Base (12)---refrigerator, desk, curtain, bookshelf, bed, table, window, cabinet, door, chair, floor, wall. \\
Novel (45)---trash can, ceiling, doorframe, object, shelf, sink, picture, backpack, couch, box, pillow, radiator, mirror, whiteboard, lamp, toilet, book, monitor, towel, tv, clothes, coffee table, office chair, nightstand, bag, dresser, toilet paper, recycling bin, kitchen cabinet, bathtub, telephone, plant, stool, keyboard, shoe, jacket, shower curtain, armchair, microwave, computer tower, bathroom vanity, kitchen counter, shower wall, paper towel dispenser, file cabinet.

\textit{ScanNet++}: Base (12)---wall, floor, door, ceiling, table, window, box, ceiling lamp, light switch, cabinet, chair, heater. \\
Novel (18)---monitor, bottle, whiteboard, office chair, doorframe, keyboard, window frame, mouse, paper, blinds, trash can, telephone, book, shelf, sink, windowsill, bag, smoke detector.

These splits reflect the long-tailed, open-set nature of indoor 3D scenes, and serve as a standard testbed for evaluating generalized few-shot capabilities. Our results are directly comparable to those of GFS-VL and its variants.

\section{Implementation Details and Reproducibility}
\label{app:impl_repro}

\paragraph{Overall protocol.}
We follow the standard two-phase generalized few-shot training protocol.
In Phase~1, the segmentation network is trained on base classes only.
In Phase~2, the model is adapted to novel classes with few-shot support annotations while being evaluated on all classes.
Our proposed components, UG-Update and UB-Filter, are activated only in Phase~2; Phase~1 follows GFS-VL~\cite{An2025}.

\paragraph{Backbone and pseudo supervision.}
We adopt Point Transformer v3 (PTv3) as the 3D backbone and use a frozen 3D vision--language model to provide dense pseudo supervision during Phase~2.
The vision--language model is used solely to produce per-point pseudo labels and the associated predictive scores needed for uncertainty estimation; it is not fine-tuned.

\paragraph{UG-Update: uncertainty-gated residual gradient update.}
UG-Update is applied after backpropagation and before each optimizer step in Phase~2.
We compute episodic uncertainty $u_{\mathrm{e}}$ from the intra-class dispersion of novel-class features under the merged support/pseudo-label supervision within the current batch (Eq.~\eqref{eq:uepi_raw}), normalized by an exponential moving average ($\beta{=}0.99$) of running mean/variance (Eq.~\eqref{eq:uepi_norm}); during the first $10$ steps the EMA statistics are still accumulating and we return $u_{\mathrm{e}}{=}0$, so the gate is essentially open ($s\approx 1$) and UG-Update reduces to a no-op until normalization is well-defined.
We then decompose the Phase-2 adaptation gradient $g$ into a base-aligned component $g_{\parallel}$ and a residual $g_{\perp}$ via projection onto a fixed base-gradient subspace (Eq.~\eqref{eq:decomp}).
The basis $B$ is constructed once at the end of Phase~1 by sampling $T{=}500$ per-batch base-task loss gradients $\nabla_{\phi}\mathcal{L}_{\mathrm{base}}(x;\phi^{*}_{\mathrm{base}})$ on the base validation stream, then applying per-parameter QR factorization and retaining the first $r$ orthonormal columns to form a rank-$r$ basis (cf.\ Eq.~\eqref{eq:basis_qr}); $B$ is then frozen throughout Phase~2.
Finally, UG-Update gates the residual update via the sigmoid coefficient $s=\sigma(-k(u_{\mathrm{e}}-\tau))$ to produce the controlled gradient $g_{\mathrm{UG}}=g_{\parallel}+s g_{\perp}$ (Eq.~\eqref{eq:gug}).
$\mathcal{P}_{\mathrm{UG}}$ comprises the PTv3 backbone parameters and the base segmentation head; the novel-class head is excluded so its parameters remain unconstrained, preserving novel-class adaptation capacity. Implementation-wise, an empty \texttt{apply\_to} list in our config is interpreted as ``apply to all parameters that have a stored basis,'' which after pretraining covers exactly $\mathcal{P}_{\mathrm{UG}}$.

\paragraph{UB-Filter: uncertainty-based pseudo-label filtering.}
UB-Filter improves the reliability of dense pseudo supervision in Phase~2 by discarding uncertain pseudo-labeled points.
Given per-point predictive probabilities from the frozen vision--language model (Eq.~\eqref{eq:vlm_prob}), we compute a point-wise uncertainty score using one of three alternatives: entropy (Eq.~\eqref{eq:upl_entropy}), margin (Eq.~\eqref{eq:upl_margin}), or max-probability (Eq.~\eqref{eq:upl_maxlogit}).
We perform class-wise quantile filtering over novel pseudo labels: for each novel class $c$, we retain the lowest-uncertainty $\rho$ fraction of points and map all others to the ignore label.
Importantly, the class-specific filtering threshold $q_c$ is not treated as a tunable hyperparameter; it is computed on-the-fly from the uncertainty distribution within class $c$ via Eq.~\eqref{eq:quantile}.
This keeps the training objective unchanged while ensuring that only high-confidence dense supervision contributes to the loss.

\paragraph{Optimization and hyperparameters.}
All experiments follow the GFS-VL~\cite{An2025} optimization setup. The only difference across UGA variants is the point-wise uncertainty definition used by UB-Filter (Entropy/Margin/Max). All key training and UGA settings are summarized in Table~\ref{tab:impl_hparams}.

\begin{table}[!htbp]
\centering
\caption{Key settings for reproducing Phase~1 pretraining and Phase~2 adaptation.}
\label{tab:impl_hparams}
\small
\setlength{\tabcolsep}{6pt}
\renewcommand{\arraystretch}{1.08}
\begin{tabular}{lcc}
\toprule
\textbf{Setting} & \textbf{ScanNet200} & \textbf{ScanNet++} \\
\midrule
\rowcolor{gray!10}\multicolumn{3}{l}{\textbf{Phase 1: base pretraining}} \\
Backbone & PTv3 & same \\
Optimizer & AdamW & same \\
Learning rate & $6\times10^{-3}$ & same \\
Weight decay & $5\times10^{-2}$ & same \\
\midrule
\rowcolor{gray!10}\multicolumn{3}{l}{\textbf{Phase 2: adaptation (UGA enabled)}} \\
Optimizer & Adam & same \\
Novel/classifier LR & $1\times10^{-3}$ & same \\
Backbone LR & $1\times10^{-4}$ & same \\
AMP & enabled & same \\
\midrule
\rowcolor{gray!10}\multicolumn{3}{l}{\textbf{UG-Update}} \\
Controlled subset $\mathcal{P}_{\mathrm{UG}}$ & PTv3 backbone $+$ base seg.\ head (novel head excluded) & same \\
Gate slope & $k=7$ & same \\
Gate threshold & $\tau=1.8$ & same \\
Uncertainty normalization & running mean/var (Eq.~\eqref{eq:uepi_norm}) & same \\
EMA momentum & $\beta=0.99$ & same \\
EMA stability constant & $\epsilon=10^{-8}$ & same \\
EMA warmup (stats accumulation) & first $10$ steps return $u_{\mathrm{e}}{=}0$ & same \\
Gradient samples for basis & $T=500$ per-batch base loss gradients (Eq.~\eqref{eq:base_grad_set}) & same \\
Basis construction & per-parameter QR factorization, retain first $r$ columns of $Q$ (Eq.~\eqref{eq:basis_qr}) & same \\
Basis rank (per-param) & $r\!=\!\min(50,\dim(\phi))$ & same \\
\midrule
\rowcolor{gray!10}\multicolumn{3}{l}{\textbf{UB-Filter}} \\
Filtering granularity & class-wise (novel only) & same \\
Retention ratio & $\rho=0.1$ & same \\
Threshold & computed $q_c$ (Eq.~\eqref{eq:quantile}) & same \\
Uncertainty definition & Entropy / Margin / Max & same \\
\bottomrule
\end{tabular}
\end{table}

\paragraph{Reproducibility notes.}
We ensure that episodic uncertainty used for UG-Update is consistent across distributed workers by synchronizing it before computing the gate coefficient.
For each independent run, the optimizer state is reset at the beginning of Phase~2 adaptation to avoid cross-run coupling via momentum statistics.

\paragraph{Compute resources.}
Experiments use 4 NVIDIA A100 GPUs (Sec.~\ref{sec:expsetup}).
In our environment, one full run through Phase~1 pretraining and Phase~2 adaptation for a single experimental setting typically requires on the order of 3--4~hours wall-clock time on this 4-GPU node (storage I/O and framework overhead dominate variance).
A representative training log measures 3~hours~48~minutes~10~seconds from loading the configuration to completion of the final aggregated evaluation when averaging over five fixed seeds (\texttt{regis1}--\texttt{regis5}; each seed fully determines initialization and episodic sampling, so every run is reproducible)---corresponding to $\approx 13690$~s ($\approx 3.80$~hours); intermediate evaluation checkpoints were spaced by roughly 33~minutes per seed on that machine.
We did not systematically log peak GPU memory; Phase~2 uses automatic mixed precision (AMP) as in Table~\ref{tab:impl_hparams}.
Hyperparameter and ablation studies in the appendix required additional compute on the same GPU class beyond these representative runs.

\paragraph{Data loading and batch sizes.}
\label{app:batch_config}
Unless otherwise overridden in code, training uses \texttt{batch\_size\_per\_gpu}${}=4$ on each of the four GPUs (global batch of $16$ samples per optimization step under standard distributed data parallel); validation uses \texttt{batch\_size\_val\_per\_gpu}${}=1$ and testing uses \texttt{batch\_size\_test\_per\_gpu}${}=1$; each GPU runs \texttt{num\_worker\_per\_gpu}${}=8$ DataLoader workers for input pipelines.

\paragraph{Multi-seed variability (representative log).}
\label{app:seed_stats}
Following Sec.~\ref{sec:expsetup}, we report means over five independent runs with fixed seeds \texttt{regis1}--\texttt{regis5}; each seed is deterministic, and the five runs differ only in the pre-specified seed value (which in turn controls initialization and episodic sampling).
Table~\ref{tab:seed_variability} reports the mean and sample standard deviation of the primary HM metric across these five runs for the reproduced baseline and our default UGA-Entropy setting. The HM gains over the reproduced baseline exceed the UGA standard deviation on all four splits.

\begin{table}[!htbp]
\centering
\caption{Five-run HM variability for the reproduced baseline and default UGA-Entropy. Entries are mean$\pm$sample std in HM mIoU (\%) over \texttt{regis1}--\texttt{regis5}.}
\label{tab:seed_variability}
\small
\setlength{\tabcolsep}{6pt}
\renewcommand{\arraystretch}{1.05}
\begin{tabular}{llc}
\toprule
Setting & Method & HM \\
\midrule
ScanNet200 5-shot & GFS-VL (Reproduced) & $42.92{\pm}0.49$ \\
ScanNet200 5-shot & UGA-Entropy & $45.65{\pm}0.48$ \\
\midrule
ScanNet200 1-shot & GFS-VL (Reproduced) & $40.39{\pm}0.75$ \\
ScanNet200 1-shot & UGA-Entropy & $43.83{\pm}0.51$ \\
\midrule
ScanNet++ 5-shot & GFS-VL (Reproduced) & $31.93{\pm}2.29$ \\
ScanNet++ 5-shot & UGA-Entropy & $36.78{\pm}1.91$ \\
\midrule
ScanNet++ 1-shot & GFS-VL (Reproduced) & $28.47{\pm}2.04$ \\
ScanNet++ 1-shot & UGA-Entropy & $31.39{\pm}1.55$ \\
\bottomrule
\end{tabular}
\end{table}

\paragraph{Licenses for existing assets.}
ScanNet~\cite{ScanNet2017} and ScanNet200~\cite{Rozenberszki2022} are used under the ScanNet Terms of Use; ScanNet++~\cite{ScanNetPP2023} is used under the ScanNet++ Terms of Use. Both require a signed agreement for academic research. The PTv3 backbone~\cite{Wu2024PTv3} is released under the MIT License; RegionPLC~\cite{Yang2024RegionPLC} is released under the Apache~2.0 License; the GFS-VL codebase~\cite{An2025} is released under the MIT License. We comply with all applicable license terms.

\section{Results on Standard ScanNet}
\label{app:scannet_split}

Following GFS-VL~\cite{An2025}, we report results on ScanNet~\cite{ScanNet2017} with 13 base and 6 novel categories (1-shot and 5-shot), using the same training protocol as in the main paper.
This ancillary setting uses a coarser class inventory and a compact base--novel partition that predates large-vocabulary generalization benchmarks; our primary experiments on ScanNet200 and ScanNet++ instead adopt wider label spaces, long-tailed base/novel splits under GFS-3DS, and scenes with heavier clutter---we therefore treat ScanNet200/++ as the more demanding evaluation and report standard ScanNet mainly for strict alignment with prior GFS-VL reporting.
Quantitative comparisons between the baseline and UGA are summarized in Table~\ref{tab:scannet_13_6}.
UGA outperforms GFS-VL by $+1.81$ HM on 1-shot and $+1.72$ HM on 5-shot, with consistent lifts on both Base and Novel, mirroring the trend observed on ScanNet200/++.
Because this split contains only $6$ novel classes (vs.\ $45$ on ScanNet200 and $18$ on ScanNet++), the fact that our gains persist here indicates that UGA's benefit does not hinge on a long-tailed novel vocabulary: episodic reliability gating and pseudo-label uncertainty pruning improve adaptation even in short-vocabulary regimes, where the main source of error is sparse geometry rather than category competition.

\begin{table}[!htbp]
\centering
\caption{ScanNet with 13 base and 6 novel classes. Results in mIoU (\%). \textbf{Bold}: best per column.}
\label{tab:scannet_13_6}
\setlength{\tabcolsep}{6pt}
\renewcommand{\arraystretch}{1.05}
\small
\begin{tabular}{ll|cccc}
\toprule
Setting & Method & B & N & A & HM \\
\midrule
1-shot & GFS-VL & 78.56 & 49.72 & 69.45 & 60.88 \\
1-shot & \textbf{UGA} & \textbf{79.84} & \textbf{51.60} & \textbf{70.92} & \textbf{62.69} \\
\midrule
5-shot & GFS-VL & 78.30 & 51.22 & 69.75 & 61.91 \\
5-shot & \textbf{UGA} & \textbf{79.22} & \textbf{53.17} & \textbf{70.99} & \textbf{63.63} \\
\bottomrule
\end{tabular}
\end{table}

\section{Generality Across Backbones and VLMs}
\label{app:generality_tables}

\noindent\textbf{Backbone generality.}
Table~\ref{tab:generality_backbone} shows that UGA consistently improves over GFS-VL under SCN~\cite{Graham2018SCN}, with gains of +1.83 / +2.42 / +2.30 / +2.51 in B/N/A/HM.
The transfer is mechanistic rather than tuned: both components of UGA operate on backbone outputs generically---UG-Update builds its base-gradient subspace $B$ from whatever Phase~1 gradients the backbone produces and decomposes every Phase~2 gradient through the same projector, while UB-Filter only consumes per-point VLM probabilities and is agnostic to how the backbone encodes geometry.
The larger gain on Novel (+2.42) than Base (+1.83) is consistent with UG-Update's intended role: attenuating residual updates during unreliable episodes mostly helps novel-class adaptation, where sparse supports are the dominant noise source.

\noindent\textbf{VLM generality.}
Table~\ref{tab:generality_vlm} shows that UGA improves performance under both RegionPLC and OpenScene~\cite{Peng2023}.
OpenScene is the considerably weaker VLM on this split (baseline HM $30.87$ vs.\ RegionPLC's $43.12$), yet UGA still delivers $+2.78$ HM on top of it, compared with $+2.53$ on the stronger RegionPLC.
This is consistent with the design logic of UB-Filter: as VLM reliability drops, the heavy tail of high-uncertainty pseudo-labels enlarges, so class-wise quantile pruning has more noise to remove and its marginal contribution does not shrink with VLM quality.
UG-Update's dispersion-based gate is computed from backbone features alone and therefore inherits none of the VLM's calibration drift, which further explains why the Novel-class lift (+2.20) persists even under the weaker VLM.

\begin{table*}[!htbp]
\centering
\begingroup
\footnotesize
\setlength{\tabcolsep}{5pt}%
\renewcommand{\arraystretch}{1.06}%
\captionsetup{font=footnotesize,labelfont={footnotesize,bf},textfont=footnotesize,justification=centering,singlelinecheck=false,margin=0pt}%
\begin{minipage}[t]{0.48\textwidth}
\captionsetup{width=\linewidth}
\centering
\captionof{table}{Backbone generality. Best: \textbf{bold}; second: \underline{underline}.}\par
\label{tab:generality_backbone}
\resizebox{\linewidth}{!}{%
\begin{tabular}{lcccc}
\toprule
Method & B & N & A & HM \\
\midrule
Baseline (SCN) & \second{61.79} & \second{32.01} & \second{38.28} & \second{42.17} \\
UGA (SCN)   & \best{63.62} & \best{34.43} & \best{40.58} & \best{44.68} \\
\bottomrule
\end{tabular}%
}
\end{minipage}
\hfill
\begin{minipage}[t]{0.48\textwidth}
\captionsetup{width=\linewidth}
\centering
\captionof{table}{VLM generality. Best: \textbf{bold}; second: \underline{underline}.}\par
\label{tab:generality_vlm}
\resizebox{\linewidth}{!}{%
\begin{tabular}{llcccc}
\toprule
VLM & Method & B & N & A & HM \\
\midrule
\multirow{2}{*}{RegionPLC}
& GFS-VL & \second{67.57} & \second{31.67} & \second{39.23} & \second{43.12} \\
& UGA   & \best{69.36} & \best{34.02} & \best{41.46} & \best{45.65} \\
\midrule
\multirow{2}{*}{OpenScene}
& GFS-VL & \second{68.41} & \second{19.93} & \second{30.14} & \second{30.87} \\
& UGA   & \best{70.22} & \best{22.13} & \best{32.25} & \best{33.65} \\
\bottomrule
\end{tabular}%
}
\end{minipage}
\endgroup
\end{table*}

\section{UG-Update Hyperparameter Study}
\label{app:hyper_sens_ugupdate}
\label{app:subspace_ablation}

\begin{figure}[ht]
  \centering
  \includegraphics[width=0.9\columnwidth]{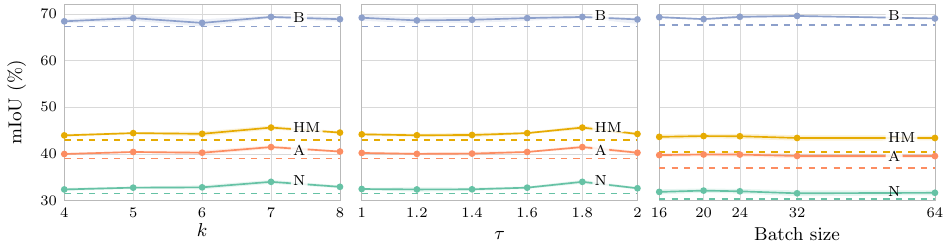}
  \caption{Hyper-parameter sensitivity on ScanNet200: $k$ (left, 5-shot), $\tau$ (middle, 5-shot), and Phase-2 batch size (right, 1-shot). Shaded regions denote 95\% confidence intervals over 5 runs. Dashed lines indicate the GFS-VL~\cite{An2025} HM baselines.}
  \label{fig:sens_k_tau_scannet200_5shot}

  \vspace{1.25\baselineskip}
  \centering
  \renewcommand{\arraystretch}{1.05}
  \setlength{\tabcolsep}{4pt}
  \begin{minipage}[t]{0.49\textwidth}
  \centering
  \small
  \captionof{table}{Subspace rank $r$ ($T{=}500$). ScanNet200 5-shot, mIoU (\%). $\dagger$: default. \textbf{Bold}: best per column.}
  \label{tab:subspace_ablation_K}
  \begin{tabular}{c|cccc}
  \toprule
  $r$ & B & N & A & HM \\
  \midrule
  10  & 68.64 & 33.58 & 40.96 & 45.10 \\
  25  & 69.38 & 34.33 & 41.71 & 45.93 \\
  50$^\dagger$ & 69.36 & 34.02 & 41.46 & 45.65 \\
  100 & 68.78 & 33.96 & 41.29 & 45.47 \\
  \textbf{200} & \textbf{69.62} & \textbf{34.36} & \textbf{41.78} & \textbf{46.01} \\
  \bottomrule
  \end{tabular}
  \end{minipage}
  \hfill
  \begin{minipage}[t]{0.49\textwidth}
  \centering
  \small
  \captionof{table}{Stored gradients $T$ ($r{=}50$). ScanNet200 5-shot, mIoU (\%). $\dagger$: default. \textbf{Bold}: best per column.}
  \label{tab:subspace_ablation_T}
  \begin{tabular}{c|cccc}
  \toprule
  $T$ & B & N & A & HM \\
  \midrule
  100  & 69.30 & 33.87 & 41.33 & 45.50 \\
  200  & 69.23 & 33.71 & 41.19 & 45.34 \\
  500$^\dagger$ & \textbf{69.36} & 34.02 & \textbf{41.46} & \textbf{45.65} \\
  1000 & 68.70 & \textbf{34.11} & 41.39 & 45.58 \\
  \bottomrule
  \end{tabular}
  \end{minipage}
\end{figure}

\noindent\textbf{Sensitivity to $k$, $\tau$, and Phase~2 batch size.}
We study the sensitivity of UG-Update to the gate $s=\sigma(-k(u_{\mathrm{e}}-\tau))$ in Fig.~\ref{fig:sens_k_tau_scannet200_5shot} (left and middle), including the slope $k$ and threshold $\tau$.
Sweeping $k\in[4,8]$ results in only mild fluctuations, suggesting robustness: small $k$ may under-suppress $g_{\perp}$ in high-uncertainty episodes, while overly large $k$ can be too conservative.
Overall, $k{=}7$ yields the best trade-off, improving over the baseline.
Similarly, sweeping $\tau\in[1.0,2.0]$ shows that smaller $\tau$ triggers earlier gating and reduces novel adaptation, whereas larger $\tau$ weakens suppression on unreliable episodes.
The best performance is achieved at $\tau{=}1.8$ (matching the optimal $k$), and the narrow confidence intervals indicate stable behavior.
The right panel of Fig.~\ref{fig:sens_k_tau_scannet200_5shot} reports Phase-2 batch size on ScanNet200 (1-shot), with 95\% confidence intervals over five runs. Performance is generally stable across batch sizes from 16 to 64, indicating UGA is not highly sensitive to this choice. Batch sizes of 20 and 24 offer the best overall balance, while larger batches slightly favor base preservation at the cost of plasticity.

\noindent\textbf{Gradient subspace ($T$, $r$).}
UG-Update builds a rank-$r$ orthonormal basis from $T$ Phase~1 gradient samples (Sec.~\ref{subsec:ug_update}).
Table~\ref{tab:subspace_ablation_K} varies $r$ with fixed $T{=}500$; Table~\ref{tab:subspace_ablation_T} varies $T$ with fixed $r{=}50$.
All runs use ScanNet200 5-shot; the published GFS-VL~\cite{An2025} HM is 43.12\%.

Moderate ranks $r\in[25,200]$ yield HM between 45.47 and 46.01, all well above GFS-VL (43.12\% HM); $r{=}10$ under-covers the base gradient manifold.
Varying $T\in[100,1000]$ changes HM by at most $\pm0.31$ percentage points, indicating a stable subspace estimate.

\section{Additional Qualitative Results}
\label{app:more_vis}

\begin{figure*}[!htbp]
  \centering
  \includegraphics[width=0.95\textwidth]{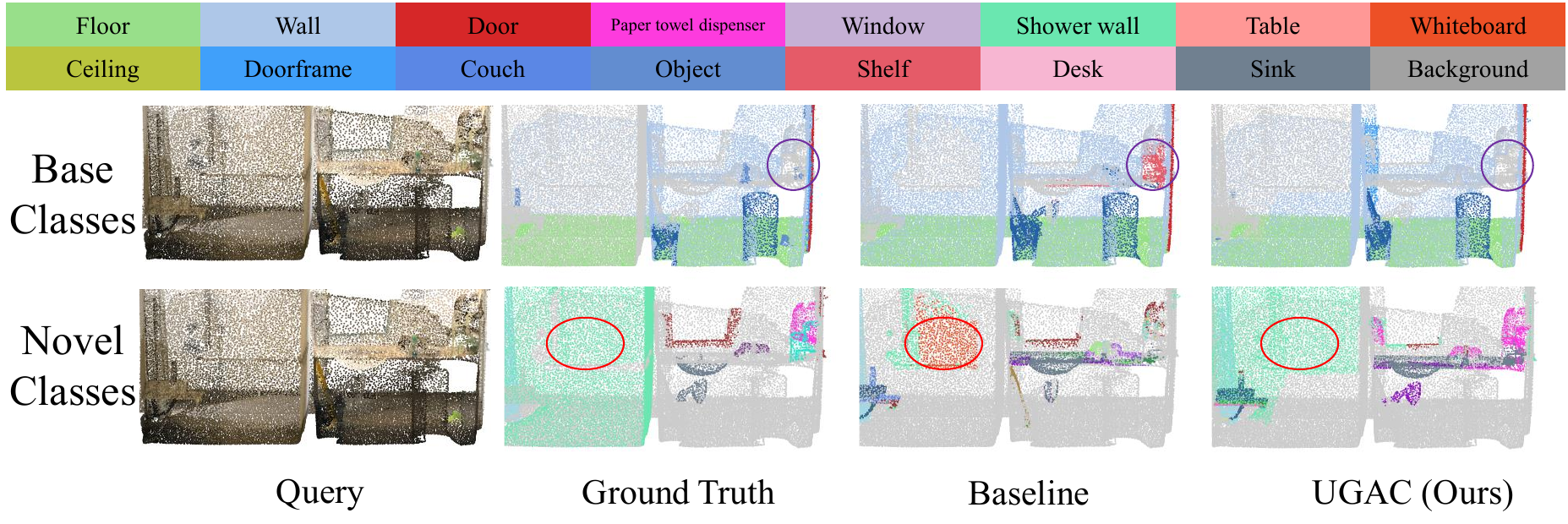}
  \caption{Zoom-in qualitative comparison with base and novel predictions visualized separately; purple/red circles mark base/novel errors.}
  \label{fig:more_vis_zoom}
\end{figure*}

Figure~\ref{fig:more_vis_zoom} zooms into the second scene of Figure~\ref{fig:main_qual} with base/novel rows shown separately.
UGA better preserves base-class predictions and more accurately segments novel classes (e.g., correctly identifying \textit{shower wall} instead of \textit{whiteboard}).
These improvements align with the quantitative results in Table~\ref{tab:main_results}.


\clearpage
\section*{NeurIPS Paper Checklist}

\begin{enumerate}

\item {\bf Claims}
    \item[] Question: Do the main claims made in the abstract and introduction accurately reflect the paper's contributions and scope?
    \item[] Answer: \answerYes{}
    \item[] Justification: Abstract and Introduction state UGA (\textbf{UG-Update}+\textbf{UB-Filter}), update-level uncertainty gating in GFS-3DS, and SoTA results on ScanNet200/++. Sec.~3 (Methodology) and Sec.~4 (Experiments) match those claims.
    \item[] Guidelines:
    \begin{itemize}
        \item The answer \answerNA{} means that the abstract and introduction do not include the claims made in the paper.
        \item The abstract and/or introduction should clearly state the claims made, including the contributions made in the paper and important assumptions and limitations. A \answerNo{} or \answerNA{} answer to this question will not be perceived well by the reviewers. 
        \item The claims made should match theoretical and experimental results, and reflect how much the results can be expected to generalize to other settings. 
        \item It is fine to include aspirational goals as motivation as long as it is clear that these goals are not attained by the paper. 
    \end{itemize}

\item {\bf Limitations}
    \item[] Question: Does the paper discuss the limitations of the work performed by the authors?
    \item[] Answer: \answerYes{}
    \item[] Justification: Conclusion discusses indoor-only evaluation, Phase~1 subspace retuning for other sensors, limits of UB-Filter on confidently wrong pseudo labels and fixed base-gradient subspace for UG-Update, plus deployment-oriented privacy/data-governance considerations.
    \item[] Guidelines:
    \begin{itemize}
        \item The answer \answerNA{} means that the paper has no limitation while the answer \answerNo{} means that the paper has limitations, but those are not discussed in the paper. 
        \item The authors are encouraged to create a separate ``Limitations'' section in their paper.
        \item The paper should point out any strong assumptions and how robust the results are to violations of these assumptions (e.g., independence assumptions, noiseless settings, model well-specification, asymptotic approximations only holding locally). The authors should reflect on how these assumptions might be violated in practice and what the implications would be.
        \item The authors should reflect on the scope of the claims made, e.g., if the approach was only tested on a few datasets or with a few runs. In general, empirical results often depend on implicit assumptions, which should be articulated.
        \item The authors should reflect on the factors that influence the performance of the approach. For example, a facial recognition algorithm may perform poorly when image resolution is low or images are taken in low lighting. Or a speech-to-text system might not be used reliably to provide closed captions for online lectures because it fails to handle technical jargon.
        \item The authors should discuss the computational efficiency of the proposed algorithms and how they scale with dataset size.
        \item If applicable, the authors should discuss possible limitations of their approach to address problems of privacy and fairness.
        \item While the authors might fear that complete honesty about limitations might be used by reviewers as grounds for rejection, a worse outcome might be that reviewers discover limitations that aren't acknowledged in the paper. The authors should use their best judgment and recognize that individual actions in favor of transparency play an important role in developing norms that preserve the integrity of the community. Reviewers will be specifically instructed to not penalize honesty concerning limitations.
    \end{itemize}

\item {\bf Theory assumptions and proofs}
    \item[] Question: For each theoretical result, does the paper provide the full set of assumptions and a complete (and correct) proof?
    \item[] Answer: \answerNA{}
    \item[] Justification: The contribution is primarily empirical and algorithmic; there are no numbered theorems/lemmas with full proof obligations as the paper's core results.
    \item[] Guidelines:
    \begin{itemize}
        \item The answer \answerNA{} means that the paper does not include theoretical results. 
        \item All the theorems, formulas, and proofs in the paper should be numbered and cross-referenced.
        \item All assumptions should be clearly stated or referenced in the statement of any theorems.
        \item The proofs can either appear in the main paper or the supplemental material, but if they appear in the supplemental material, the authors are encouraged to provide a short proof sketch to provide intuition. 
        \item Inversely, any informal proof provided in the core of the paper should be complemented by formal proofs provided in appendix or supplemental material.
        \item Theorems and Lemmas that the proof relies upon should be properly referenced. 
    \end{itemize}

    \item {\bf Experimental result reproducibility}
    \item[] Question: Does the paper fully disclose all the information needed to reproduce the main experimental results of the paper to the extent that it affects the main claims and/or conclusions of the paper (regardless of whether the code and data are provided or not)?
    \item[] Answer: \answerYes{}
    \item[] Justification: Sec.~4.1 states benchmarks, protocol, backbone/VLM, hardware, and averaging over five runs; Appendix~\ref{app:impl_repro} summarizes phases, objectives, UG-Update/UB-Filter settings, and Table~\ref{tab:impl_hparams} lists key hyperparameters; Appendix~\ref{app:batch_config} documents distributed batch sizes and DataLoader workers; Appendix~\ref{app:seed_stats} reports representative multi-seed variability.
    \item[] Guidelines:
    \begin{itemize}
        \item The answer \answerNA{} means that the paper does not include experiments.
        \item If the paper includes experiments, a \answerNo{} answer to this question will not be perceived well by the reviewers: Making the paper reproducible is important, regardless of whether the code and data are provided or not.
        \item If the contribution is a dataset and\slash or model, the authors should describe the steps taken to make their results reproducible or verifiable. 
        \item Depending on the contribution, reproducibility can be accomplished in various ways. For example, if the contribution is a novel architecture, describing the architecture fully might suffice, or if the contribution is a specific model and empirical evaluation, it may be necessary to either make it possible for others to replicate the model with the same dataset, or provide access to the model. In general. releasing code and data is often one good way to accomplish this, but reproducibility can also be provided via detailed instructions for how to replicate the results, access to a hosted model (e.g., in the case of a large language model), releasing of a model checkpoint, or other means that are appropriate to the research performed.
        \item While NeurIPS does not require releasing code, the conference does require all submissions to provide some reasonable avenue for reproducibility, which may depend on the nature of the contribution. For example
        \begin{enumerate}
            \item If the contribution is primarily a new algorithm, the paper should make it clear how to reproduce that algorithm.
            \item If the contribution is primarily a new model architecture, the paper should describe the architecture clearly and fully.
            \item If the contribution is a new model (e.g., a large language model), then there should either be a way to access this model for reproducing the results or a way to reproduce the model (e.g., with an open-source dataset or instructions for how to construct the dataset).
            \item We recognize that reproducibility may be tricky in some cases, in which case authors are welcome to describe the particular way they provide for reproducibility. In the case of closed-source models, it may be that access to the model is limited in some way (e.g., to registered users), but it should be possible for other researchers to have some path to reproducing or verifying the results.
        \end{enumerate}
    \end{itemize}

\item {\bf Open access to data and code}
    \item[] Question: Does the paper provide open access to the data and code, with sufficient instructions to faithfully reproduce the main experimental results, as described in supplemental material?
    \item[] Answer: \answerYes{}
    \item[] Justification: We release a code repository at \url{https://anonymous.4open.science/r/UGA/} containing the full implementation, training/evaluation scripts, configuration files, and run instructions consistent with Appendix~\ref{app:impl_repro}. Datasets are the public ScanNet200/ScanNet++ benchmarks accessed as in the cited original works.
    \item[] Guidelines:
    \begin{itemize}
        \item The answer \answerNA{} means that paper does not include experiments requiring code.
        \item Please see the NeurIPS code and data submission guidelines (\url{https://neurips.cc/public/guides/CodeSubmissionPolicy}) for more details.
        \item While we encourage the release of code and data, we understand that this might not be possible, so \answerNo{} is an acceptable answer. Papers cannot be rejected simply for not including code, unless this is central to the contribution (e.g., for a new open-source benchmark).
        \item The instructions should contain the exact command and environment needed to run to reproduce the results. See the NeurIPS code and data submission guidelines (\url{https://neurips.cc/public/guides/CodeSubmissionPolicy}) for more details.
        \item The authors should provide instructions on data access and preparation, including how to access the raw data, preprocessed data, intermediate data, and generated data, etc.
        \item The authors should provide scripts to reproduce all experimental results for the new proposed method and baselines. If only a subset of experiments are reproducible, they should state which ones are omitted from the script and why.
        \item At submission time, to preserve anonymity, the authors should release anonymized versions (if applicable).
        \item Providing as much information as possible in supplemental material (appended to the paper) is recommended, but including URLs to data and code is permitted.
    \end{itemize}

\item {\bf Experimental setting/details}
    \item[] Question: Does the paper specify all the training and test details (e.g., data splits, hyperparameters, how they were chosen, type of optimizer) necessary to understand the results?
    \item[] Answer: \answerYes{}
    \item[] Justification: Sec.~4.1 covers protocol, backbone, frozen VLM, optimizer scale (Appendix~\ref{app:impl_repro}, Table~\ref{tab:impl_hparams}); Appendix~\ref{app:benchmark} documents splits; Appendix~\ref{app:batch_config} specifies training/validation/testing batch sizes and worker counts under DDP; Appendix~\ref{app:hyper_sens_ugupdate} discusses sensitivity for selected hyperparameters.
    \item[] Guidelines:
    \begin{itemize}
        \item The answer \answerNA{} means that the paper does not include experiments.
        \item The experimental setting should be presented in the core of the paper to a level of detail that is necessary to appreciate the results and make sense of them.
        \item The full details can be provided either with the code, in appendix, or as supplemental material.
    \end{itemize}

\item {\bf Experiment statistical significance}
    \item[] Question: Does the paper report error bars suitably and correctly defined or other appropriate information about the statistical significance of the experiments?
    \item[] Answer: \answerYes{}
    \item[] Justification: Main tables report means over five fixed seeds \texttt{regis1}--\texttt{regis5} (Sec.~4.1); each seed is deterministic, so the reported variability reflects dispersion across these pre-specified registration seeds rather than stochastic re-runs. Appendix~\ref{app:seed_stats} reports HM mean$\pm$sample standard deviation for the reproduced baseline and UGA-Entropy on all ScanNet200/++ 1-/5-shot settings. Appendix~\ref{app:hyper_sens_ugupdate} reports 95\% confidence intervals over the same five fixed-seed runs for selected sensitivity curves (Fig.~\ref{fig:sens_k_tau_scannet200_5shot}).
    \item[] Guidelines:
    \begin{itemize}
        \item The answer \answerNA{} means that the paper does not include experiments.
        \item The authors should answer \answerYes{} if the results are accompanied by error bars, confidence intervals, or statistical significance tests, at least for the experiments that support the main claims of the paper.
        \item The factors of variability that the error bars are capturing should be clearly stated (for example, train/test split, initialization, random drawing of some parameter, or overall run with given experimental conditions).
        \item The method for calculating the error bars should be explained (closed form formula, call to a library function, bootstrap, etc.)
        \item The assumptions made should be given (e.g., Normally distributed errors).
        \item It should be clear whether the error bar is the standard deviation or the standard error of the mean.
        \item It is OK to report 1-sigma error bars, but one should state it. The authors should preferably report a 2-sigma error bar than state that they have a 96\% CI, if the hypothesis of Normality of errors is not verified.
        \item For asymmetric distributions, the authors should be careful not to show in tables or figures symmetric error bars that would yield results that are out of range (e.g., negative error rates).
        \item If error bars are reported in tables or plots, the authors should explain in the text how they were calculated and reference the corresponding figures or tables in the text.
    \end{itemize}

\item {\bf Experiments compute resources}
    \item[] Question: For each experiment, does the paper provide sufficient information on the computer resources (type of compute workers, memory, time of execution) needed to reproduce the experiments?
    \item[] Answer: \answerYes{}
    \item[] Justification: Sec.~4.1 specifies 4$\times$ NVIDIA A100 GPUs; Appendix~\ref{app:impl_repro} (\textit{Compute resources}) gives both an order-of-magnitude wall-clock range (\textbf{3--4~hours}) for Phase~1+Phase~2 on that 4-GPU node and a representative logged duration from configuration load through final aggregated evaluation across five seeds (\textbf{3~h~48~m~10~s}, $\approx$\textbf{13690~s} / \textbf{3.80~hours}), with intermediate evaluations spaced by $\approx$\textbf{33~minutes}; it notes AMP and states that peak GPU memory was not systematically logged. Appendix studies used the same GPU class with additional runtime.
    \item[] Guidelines:
    \begin{itemize}
        \item The answer \answerNA{} means that the paper does not include experiments.
        \item The paper should indicate the type of compute workers CPU or GPU, internal cluster, or cloud provider, including relevant memory and storage.
        \item The paper should provide the amount of compute required for each of the individual experimental runs as well as estimate the total compute. 
        \item The paper should disclose whether the full research project required more compute than the experiments reported in the paper (e.g., preliminary or failed experiments that didn't make it into the paper). 
    \end{itemize}
    
\item {\bf Code of ethics}
    \item[] Question: Does the research conducted in the paper conform, in every respect, with the NeurIPS Code of Ethics \url{https://neurips.cc/public/EthicsGuidelines}?
    \item[] Answer: \answerYes{}
    \item[] Justification: Standard computational research on public indoor RGB-D benchmarks and cited models; no ethics deviations known to the authors after reviewing the NeurIPS Code of Ethics.
    \item[] Guidelines:
    \begin{itemize}
        \item The answer \answerNA{} means that the authors have not reviewed the NeurIPS Code of Ethics.
        \item If the authors answer \answerNo, they should explain the special circumstances that require a deviation from the Code of Ethics.
        \item The authors should make sure to preserve anonymity (e.g., if there is a special consideration due to laws or regulations in their jurisdiction).
    \end{itemize}

\item {\bf Broader impacts}
    \item[] Question: Does the paper discuss both potential positive societal impacts and negative societal impacts of the work performed?
    \item[] Answer: \answerYes{}
    \item[] Justification: Introduction motivates 3D perception for embodied AI and retrieval (positive pathways); Conclusion discusses deployment-related privacy and data-governance risks alongside limitations of indoor-only evaluation.
    \item[] Guidelines:
    \begin{itemize}
        \item The answer \answerNA{} means that there is no societal impact of the work performed.
        \item If the authors answer \answerNA{} or \answerNo, they should explain why their work has no societal impact or why the paper does not address societal impact.
        \item Examples of negative societal impacts include potential malicious or unintended uses (e.g., disinformation, generating fake profiles, surveillance), fairness considerations (e.g., deployment of technologies that could make decisions that unfairly impact specific groups), privacy considerations, and security considerations.
        \item The conference expects that many papers will be foundational research and not tied to particular applications, let alone deployments. However, if there is a direct path to any negative applications, the authors should point it out. For example, it is legitimate to point out that an improvement in the quality of generative models could be used to generate Deepfakes for disinformation. On the other hand, it is not needed to point out that a generic algorithm for optimizing neural networks could enable people to train models that generate Deepfakes faster.
        \item The authors should consider possible harms that could arise when the technology is being used as intended and functioning correctly, harms that could arise when the technology is being used as intended but gives incorrect results, and harms following from (intentional or unintentional) misuse of the technology.
        \item If there are negative societal impacts, the authors could also discuss possible mitigation strategies (e.g., gated release of models, providing defenses in addition to attacks, mechanisms for monitoring misuse, mechanisms to monitor how a system learns from feedback over time, improving the efficiency and accessibility of ML).
    \end{itemize}
    
\item {\bf Safeguards}
    \item[] Question: Does the paper describe safeguards that have been put in place for responsible release of data or models that have a high risk for misuse (e.g., pre-trained language models, image generators, or scraped datasets)?
    \item[] Answer: \answerNA{}
    \item[] Justification: The paper does not newly release a large scraped dataset or a general-purpose generative model; it evaluates on established benchmarks with a frozen VLM as in prior GFS-VL-style pipelines.
    \item[] Guidelines:
    \begin{itemize}
        \item The answer \answerNA{} means that the paper poses no such risks.
        \item Released models that have a high risk for misuse or dual-use should be released with necessary safeguards to allow for controlled use of the model, for example by requiring that users adhere to usage guidelines or restrictions to access the model or implementing safety filters. 
        \item Datasets that have been scraped from the Internet could pose safety risks. The authors should describe how they avoided releasing unsafe images.
        \item We recognize that providing effective safeguards is challenging, and many papers do not require this, but we encourage authors to take this into account and make a best faith effort.
    \end{itemize}

\item {\bf Licenses for existing assets}
    \item[] Question: Are the creators or original owners of assets (e.g., code, data, models), used in the paper, properly credited and are the license and terms of use explicitly mentioned and properly respected?
    \item[] Answer: \answerYes{}
    \item[] Justification: All assets are cited; Appendix~\ref{app:impl_repro} (\textit{Licenses for existing assets}) enumerates the license or terms-of-use agreement for each dataset (ScanNet/ScanNet200 Terms of Use, ScanNet++ Terms of Use), backbone (PTv3, MIT), frozen VLM (RegionPLC, Apache~2.0), and baseline code (GFS-VL, MIT).
    \item[] Guidelines:
    \begin{itemize}
        \item The answer \answerNA{} means that the paper does not use existing assets.
        \item The authors should cite the original paper that produced the code package or dataset.
        \item The authors should state which version of the asset is used and, if possible, include a URL.
        \item The name of the license (e.g., CC-BY 4.0) should be included for each asset.
        \item For scraped data from a particular source (e.g., website), the copyright and terms of service of that source should be provided.
        \item If assets are released, the license, copyright information, and terms of use in the package should be provided. For popular datasets, \url{paperswithcode.com/datasets} has curated licenses for some datasets. Their licensing guide can help determine the license of a dataset.
        \item For existing datasets that are re-packaged, both the original license and the license of the derived asset (if it has changed) should be provided.
        \item If this information is not available online, the authors are encouraged to reach out to the asset's creators.
    \end{itemize}

\item {\bf New assets}
    \item[] Question: Are new assets introduced in the paper well documented and is the documentation provided alongside the assets?
    \item[] Answer: \answerNA{}
    \item[] Justification: No new public dataset or standalone code repository is released with this submission; the method builds on existing benchmarks and frozen models.
    \item[] Guidelines:
    \begin{itemize}
        \item The answer \answerNA{} means that the paper does not release new assets.
        \item Researchers should communicate the details of the dataset\slash code\slash model as part of their submissions via structured templates. This includes details about training, license, limitations, etc. 
        \item The paper should discuss whether and how consent was obtained from people whose asset is used.
        \item At submission time, remember to anonymize your assets (if applicable). You can either create an anonymized URL or include an anonymized zip file.
    \end{itemize}

\item {\bf Crowdsourcing and research with human subjects}
    \item[] Question: For crowdsourcing experiments and research with human subjects, does the paper include the full text of instructions given to participants and screenshots, if applicable, as well as details about compensation (if any)? 
    \item[] Answer: \answerNA{}
    \item[] Justification: The study uses standard computational experiments on public datasets; no crowdsourcing or human-subjects protocols.
    \item[] Guidelines:
    \begin{itemize}
        \item The answer \answerNA{} means that the paper does not involve crowdsourcing nor research with human subjects.
        \item Including this information in the supplemental material is fine, but if the main contribution of the paper involves human subjects, then as much detail as possible should be included in the main paper. 
        \item According to the NeurIPS Code of Ethics, workers involved in data collection, curation, or other labor should be paid at least the minimum wage in the country of the data collector. 
    \end{itemize}

\item {\bf Institutional review board (IRB) approvals or equivalent for research with human subjects}
    \item[] Question: Does the paper describe potential risks incurred by study participants, whether such risks were disclosed to the subjects, and whether Institutional Review Board (IRB) approvals (or an equivalent approval/review based on the requirements of your country or institution) were obtained?
    \item[] Answer: \answerNA{}
    \item[] Justification: No human-subjects or crowdsourcing component; IRB not applicable.
    \item[] Guidelines:
    \begin{itemize}
        \item The answer \answerNA{} means that the paper does not involve crowdsourcing nor research with human subjects.
        \item Depending on the country in which research is conducted, IRB approval (or equivalent) may be required for any human subjects research. If you obtained IRB approval, you should clearly state this in the paper. 
        \item We recognize that the procedures for this may vary significantly between institutions and locations, and we expect authors to adhere to the NeurIPS Code of Ethics and the guidelines for their institution. 
        \item For initial submissions, do not include any information that would break anonymity (if applicable), such as the institution conducting the review.
    \end{itemize}

\item {\bf Declaration of LLM usage}
    \item[] Question: Does the paper describe the usage of LLMs if it is an important, original, or non-standard component of the core methods in this research? Note that if the LLM is used only for writing, editing, or formatting purposes and does \emph{not} impact the core methodology, scientific rigor, or originality of the research, declaration is not required.
    \item[] Answer: \answerNA{}
    \item[] Justification: Core novelty is uncertainty-gated adaptation (UG-Update/UB-Filter); a frozen 3D VLM supplies pseudo labels following the same family of pipelines as GFS-VL rather than new LLM-centric methodology.
    \item[] Guidelines:
    \begin{itemize}
        \item The answer \answerNA{} means that the core method development in this research does not involve LLMs as any important, original, or non-standard components.
        \item Please refer to our LLM policy in the NeurIPS handbook for what should or should not be described.
    \end{itemize}

\end{enumerate}

\end{document}